\documentclass[pmlr,twocolumn,10pt]{jmlr} % W&CP article

\usepackage{booktabs}
\usepackage{siunitx}

\usepackage[switch]{lineno}

\usepackage{blindtext}
\usepackage{amsmath}

\theorembodyfont{\upshape}
\theoremheaderfont{\scshape}
\theorempostheader{:}
\theoremsep{\newline}

\title[Visual Concept Ranking]{Visual concept ranking uncovers medical shortcuts used by large multimodal models}

\author{%
\Name{Joseph D. Janizek} \Email{jjanizek@stanford.edu}\\
\addr Department of Radiology, Stanford University
\AND
\Name{Sonnet Xu} \Email{sonnet@stanford.edu}\\
\addr Department of Computer Science, Stanford University
\AND
\Name{Junayd Lateef} \Email{junayd@stanford.edu}\\
\addr Department of Bioengineering, University of California, Berkeley
\AND
\Name{Roxana Daneshjou} \Email{roxanad@stanford.edu}\\
\addr Department of Biomedical Data Science, Department of Dermatology, Stanford University\\
}

\begin{document}

\maketitle

\begin{abstract}
Ensuring the reliability of machine learning models in safety-critical domains such as healthcare requires auditing methods that can uncover model shortcomings. 
% In addition to traditional audits like costly clinical trials or automatic multiple-choice benchmark evaluations that measure models' outputs,
We introduce a method for identifying important visual concepts within large multimodal models (LMMs) and use it to investigate the behaviors these models exhibit when prompted with medical tasks. We primarily focus on the task of classifying malignant skin lesions from clinical dermatology images, with supplemental experiments including both chest radiographs and natural images. After showing how LMMs display unexpected gaps in performance between different demographic subgroups when prompted with demonstrating examples, we apply our method, Visual Concept Ranking (VCR), to these models and prompts. VCR generates hypotheses related to different visual feature dependencies, which we are then able to validate with manual interventions. %We further demonstrate how this approach can be applied more broadly to other medical tasks.
\end{abstract}

\paragraph*{Data and Code Availability}
This paper uses the Diverse Dermatology Images (DDI) dataset \citep{daneshjou2022disparities} for the main text experiments and the CheXpert dataset \citep{rajpurkar2017chexnet} for supplemental experiments, both of which are available on the \href{https://aimi.stanford.edu/datasets/ddi-diverse-dermatology-images}{Stanford AIMI repository}. Additional supplemental experiments are conducted on the publicly available \href{https://github.com/fastai/imagenette}{Imagenette dataset}. Code is available at \url{https://github.com/jjanizek/vcr-paper}.

\paragraph*{Institutional Review Board (IRB)}
This research did not require IRB approval, as it was conducted using publicly available information that had been anonymized/de-identified to protect privacy. 

\section{Introduction}
\label{sec:intro}

\begin{figure*}[htp]
\begin{center}
%\framebox[4.0in]{$\;$}
\includegraphics[width=0.63\linewidth]{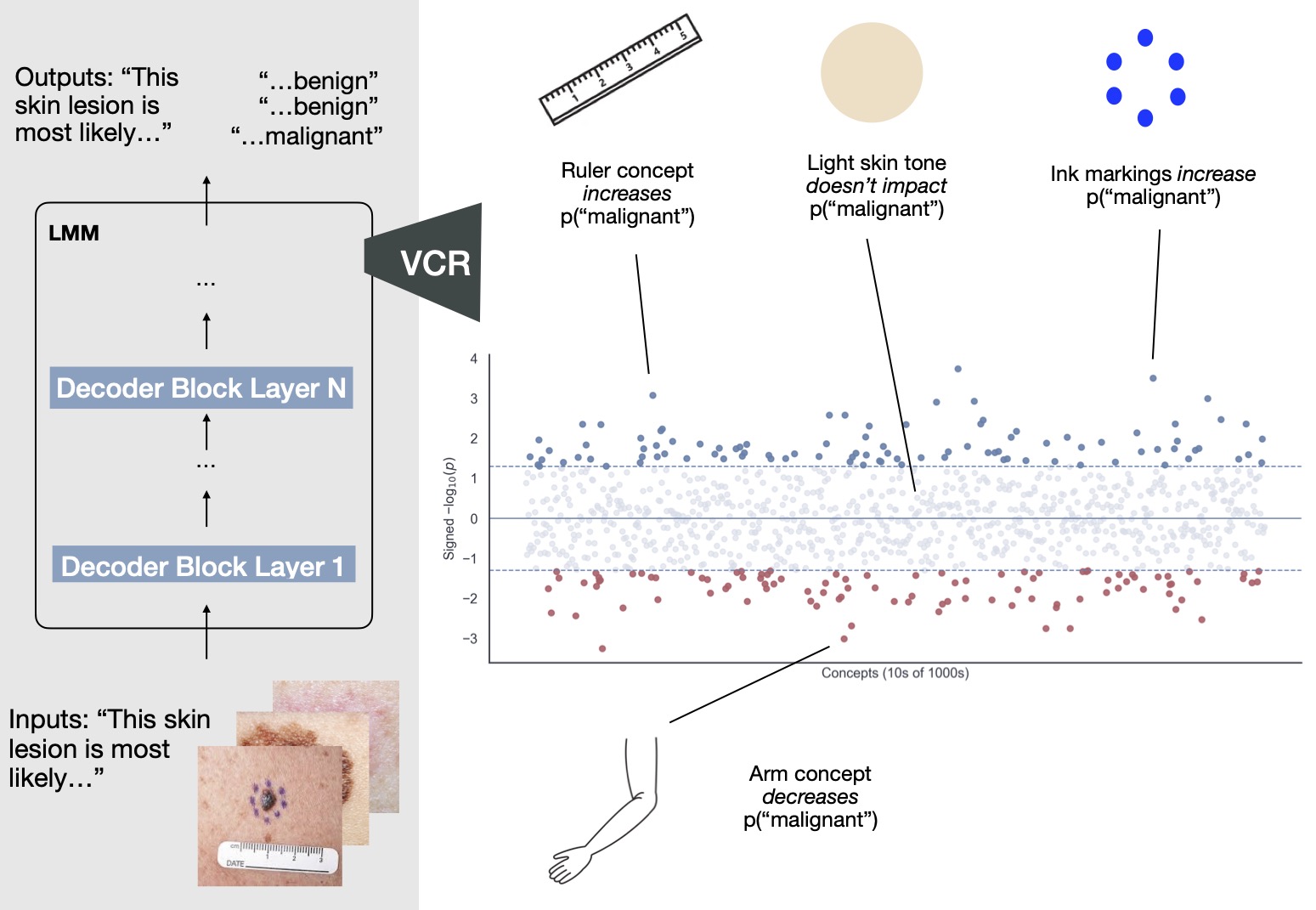}
\end{center}
\caption{Our visual concept ranking method (VCR) identifies visual concepts that have a statistically significant impact on an LMM's output for a particular task or prompt. This concept figure illustrates a hypothetical binary classification prompt for a set of clinical dermatology images, tested against tens of thousands of visual concepts.}
\label{fig:concept_method}
\end{figure*}

The last decade has seen a shift in the architectures and training recipes of the neural networks used for medical computer vision tasks. In the 2010s, convolutional neural networks trained with supervised learning first achieved state-of-the-art performance on natural image classification \citep{krizhevsky2012imagenet}, and were then applied with great success to a variety of medical fields including dermatology \citep{esteva2017dermatologist}, radiology \citep{rajpurkar2017chexnet}, and cardiology \citep{ouyang2020video}. By the early 2020s, advances in language modeling eliminated the need for laborious hand-curated supervised labels, and contrastive learning approaches like ConVIRT and CLIP allowed transformer-based vision encoders to be trained on datasets of hundreds of millions of image-caption pairs \citep{zhang2022contrastive,radford2021learning}. Finally, the current state-of-the-art for increasingly complex and open-ended tasks, such as reasoning about scientific figures \citep{laurent2024lab} or answering questions about entire CT scans \citep{hamamci2024foundation}, has been achieved by the joint training of vision encoders with large language models pre-trained on large internet-scale general text corpora, enabling the creation of large multimodal models (LMMs). These LMMs take arbitrary sequences of interleaved text and images as input, and produce a probability distribution over natural language tokens as their output \citep{wadekar2024evolution}.

While surprisingly performant, the original supervised and convolutional models discussed above also had surprising failure modes. Their performance would often degrade to the point of being useless when tested in novel hospitals outside of their development environment \citep{yu2022external}. A substantial body of AI interpretability was devoted to uncovering the mechanisms behind these failures of generalization. Early work used saliency maps to show a propensity of models to look at hospital-specific factors like laterality markers in chest radiographs \citep{zech2018variable}. While saliency maps can identify spatially-localized dependencies such as a text marker in the corner of an image, they are quite limited in expressing more abstract concepts. For instance, it is difficult to localize concepts like ``the overall brightness of an image'' to a particular set of pixels. Additionally, even if saliency can be localized to a particular object, per-pixel importance values cannot reveal what visual characteristics of that object are important to the model. Hence, further approaches have been developed and applied, such as using generative neural networks to create counterfactual images \citep{singla2019explanation,degrave2021ai}, or using linear probes to test model dependencies on concepts \citep{kim2018interpretability}.

Our work specifically focuses on applying this latter sort of conceptual interpretability to LMMs. This is of particular interest, as LMMs have already demonstrated a variety of concerning and often difficult-to-explain behaviors for medical applications. For instance, \cite{xu2025biasicl} showed that when commercial LMMs like GPT-4o are prompted to predict malignant skin lesions with an In-Context Learning (ICL) prompt, they may start to perform substantially better on one demographic subgroup at the expense of another. What the models are ``looking at'' when this occurs is currently unclear. 

To identify the visual conceptual dependencies of LMMs, we propose using a method we call Visual Concept Ranking (VCR). For a given model and a given task, our method produces a list of the most statistically significant visual concepts (\figureref{fig:concept_method} and Supplemental \figureref{fig:visual-algo}). We first validate the efficacy of this approach using synthetic data, demonstrating how our method is more robust to domain shift than prior methods. Then, focusing on the task of classification of malignant skin lesions, we use VCR to generate testable hypotheses about LMMs' behavior. These hypotheses are finally validated with manual image interventions.

\section{Method: Visual Concept Ranking (VCR)}

Our goal is to identify the most important visual concepts used by LMMs to complete specific tasks. At a high level, our method works by using a vision-language model (VLM) to automatically label a probe set of images with concept labels. These images can then be used to create concept activation vectors \citep{kim2018interpretability} for arbitrarily large numbers of concepts. Finally, the model's sensitivity to perturbation along those directions is tested to rank the importance of the concepts. 

To further clarify our terminology here, when we say LMMs in this paper, we mean models like OpenFlamingo \citep{awadalla2023openflamingo} which have been trained to produce text when prompted with text and images. When we refer to VLMs, we refer to models like CLIP \citep{radford2021learning}, which comprise an image encoder and text encoder that have been trained to map images and texts to the same embedding space. The LMM is the target model we would like to explain, and the VLM is used as an explainer model.

A task is defined by a prompt with an expected set of completions. For instance, binary classification of malignant skin lesions can be posed as a prompt (``User:[image]Is this lesion benign or malignant? Assistant: The lesion is'') and the length-normalized log-probabilities of the completions ``benign'' or ``malignant'' can be used as the binary class probabilities. 

We assume that the user has a set of unlabeled images for which they would like to investigate the model's behavior: $ \mathcal{I} \;=\; \{\,I_1,\, I_2,\, \dots,\, I_N\}$. We also assume that the user has a set of concepts they would like to test: $ \mathcal{C} \;=\; \{\,C_1,\, C_2,\, \dots,\, C_M\},$ where each \(C_i\) is a string describing a concept. In this paper we use large concept sets, such as the 20,000 most common English words \citep{google10000english}, augmented with domain specific concept lists \citep{daneshjou2022skincon}, to enable ``hypothesis-free'' concept testing, on analogy to genome-wide association screens. The ability to test very large sets of concepts is a key feature of our method, enabling the discovery of unexpected relationships between concepts and outcomes. Rather than testing a small set of preconceived hypotheses, our approach generates novel hypotheses about model mechanisms that can then be validated with more difficult and time-intensive manual interventions.

The following subsections detail the steps of our method. For a conceptual illustration of this method outline, see Supplemental \figureref{fig:visual-algo}.

\subsection{Steps of VCR Algorithm}

\begin{enumerate}
    \item \textbf{Generating concept labels for a probe set of images using a VLM:} Given an image set \( \mathcal{I} \) and a concept set \( \mathcal{C} \), we use a pre-trained vision-language model to compute a \textit{concept label} for each image–concept pair. For each image \(I_i\) and concept \(C_k\), we calculate the $\ell_2$-normalized cosine similarities between the embedding from the VLM's vision encoder, $\phi_v(I_i)$, and the embedding from the VLM's text encoder, $\phi_t(C_k)$:
\begin{equation}
    Y_{i,k} = \frac{\phi_v(I_i)^T \phi_t(C_k)}{\|\phi_v(I_i)\|\|\phi_t(C_k)\|},
\end{equation}
producing a normalized concept label matrix $Y \in \mathbb{R}^{N \times K}$. Intuitively, each entry \( Y_{i,k} \) quantifies the extent to which concept \( C_k \) is visually represented in image \( I_i \). While we use OpenCLIP \citep{ilharco2021openclip} in our experiments, we note that any pretrained VLM that has been trained to map images and texts to the same embedding space could theoretically be used here.

    \item \textbf{Learning concept activation vectors from concept labels and model activations:} 
    
    To interpret a target LMM ($f$), we extract the model's activations from a specific layer (\( \ell \)) when prompted with each image in the probe set.  We use the activations corresponding to the \textit{last token position} from the template text \citep{zou2023representation}. This yields an activation matrix \( A \in \mathbb{R}^{N \times D} \), where each row \( a_i = f_\ell(I_i)[-1] \) is the \( D \)-dimensional residual stream activation vector at layer \( \ell \) for the final token for the prompt including an image \( I_i \). For each experiment, all components of the prompt (e.g., demonstrations in the ICL setting) are held fixed except for the final image query, and consequently the model's dependence on both the rest of the prompt and the underlying model parameters is omitted from the functional notation.

    Using the concept labels \( Y \) as targets and the activation matrix \( A \) as features, we train a linear regression model for each concept \( k \in \mathcal{C} \), using \( A \) to predict the corresponding column \( Y_{\cdot,k} \):
\begin{equation}
\mathbf{\hat{w}}_k = \arg\min_{\mathbf{w}_k \in \mathbb{R}^{D}} \| A\mathbf{w}_k - \mathbf{y}_k \|_2^2 + \lambda \|\mathbf{w}_k\|_2^2.
\end{equation}
The normalized version of the vector $\mathbf{\hat{w}}_k$ defines a ``concept activation vector'' (CAV) \citep{kim2018interpretability}, which represents the unit-length direction in activation space corresponding to concept \(C_k\).
 
We use an L2 regularized model to ensure a unique solution when $D$ (order of magnitude 1000s of latents) exceeds the number of probe images $N$ (order of magnitude hundreds of samples), and to allow a closed-form solution for computational efficiency.

Given that our method is designed to test tens of thousands of concepts, this computational efficiency is essential for practical applicability.

    \item \textbf{Measuring model's sensitivity to concept vectors:} We assess the \textit{importance of each concept} to the model's output for the task of interest via directional derivatives. For classification tasks (e.g., predicting whether a skin lesion is malignant), we define a \textit{task score} as the length-normalized log-probability of the target class:
\begin{equation}
S_i = \frac{1}{L} \sum_{t=1}^{L} \log \mathbb{P}_f(\text{``malignant''}^{(t)} \mid I_i),
\end{equation}
where $L$ is the sequence length and the sum is over token positions in the target class label. 

The \textit{sensitivity} of concept \( k \) is then the directional derivative along the normalized concept vector:
\begin{equation}
\psi_k = \frac{1}{N} \sum_{i=1}^N \left\langle \nabla_{a_i} S(a_i), \, \mathbf{v}_k \right\rangle,
\end{equation}
where \( \nabla_{a_i} S(a_i) \) is the gradient of the task score with respect to the activations \( a_i \), and the sum is taken over all images in the probe set to produce a single global importance score for each concept.

\item \textbf{Significance Testing}

To assess the statistical significance of each concept's global importance score ($\psi_{k}$), we repeat our VCR process using bootstrap resampled versions of the image probing set. For each of the $K$ concepts tested, we perform a two-sided one-sample t-test comparing the distribution of the concept's global importance scores across repeated runs against the null hypothesis of zero importance. To conservatively correct for multiple comparisons across all $K$ concepts, we apply a Bonferroni correction \citep{bonferroni1936teoria}, rejecting null hypotheses only at a threshold of $p < 0.05 / K$.

\end{enumerate}

\subsection{Related Work}

A recent paper that was highly influential to our VCR method was another method for model auditing using VLMs, called MA-MONET \citep{kim2024transparent}. Like VCR, the MA-MONET method proposes using VLMs to label a probe set of images which can then be used to audit model behaviors. Unlike VCR, MA-MONET looks at correlations between visual features and model outputs rather than using model internals (i.e. activations and gradients) to directly interpret causal mechanisms. While this has some advantages compared to our method\footnote{MA-MONET can be used to audit API-access-only models where users don't have access to model internals}, we hypothesized that it may cause MA-MONET to behave counterintuitively in some frequently encountered medical settings.

To illustrate this point, consider two visual features that may be used to predict pneumonia from a chest radiograph: (1) the presence of opacities in the lung fields, and (2) a potential ``shortcut'' like a hospital-specific laterality marker in the corner of an image \citep{zech2018variable,degrave2021ai}. One can imagine a well-trained LMM that looks at lung opacities but \textit{not the shortcut}, as a result of the specific data used in its training. Now, say that a correlational approach like MA-MONET is used to audit this LMM. The user, who does not necessarily have access to the full training corpus of the model\footnote{Frequently the case with models with \textit{open-weights} but not \textit{open-source}, like LLaMa}, collects a probe set of images from their own hospital, where there is a very strong correlation between the laterality marker and pneumonia. In this case, a correlational approach like MA-MONET will notice that there is a strong association between the model's prediction of pneumonia and the presence of this shortcut feature, and report it as an important visual conceptual association, despite the fact that the model \textit{does not depend on this visual feature}, and would not be impacted by adding or removing it. Given the frequency of potential shortcuts in medical datasets, we wanted to develop a method that would report \textit{causal} dependencies on visual features (see \figureref{fig:vcr-not-fooled}).

A second, very closely related prior work is a recently published paper on Language-Guided CAVs (LG-CAV) \citep{huang2024lg}. Like VCR, LG-CAV also leverages VLMs to generate supervision signals for concept activation vectors. Our method differs in two key ways. First, LG-CAV is designed for image classification models trained with supervised learning and fixed class logits, whereas VCR is adapted for LMMs that produce probability distributions over natural language tokens (see step 3 of our algorithm outline, where a natural language ``task score” is defined). Second, LG-CAV presumes access to labeled examples of concept classes (used in its deviation sample reweighting module and classification loss), while VCR requires no ground-truth concept labels. In this sense, VCR can be viewed as an adaptation of LG-CAV tailored to LMMs and the less-structured data settings in which they are typically audited. For additional related prior work, see discussion in Appendix \ref{apd:rel_work}.

\section{Results}

\subsection{Synthetic data benchmarks}

\begin{figure}[t]
\floatconts
  {fig:synthetic-data-main}
  {\caption{\textbf{Left}, Example images from the synthetic datasets. \textbf{Right}, For two synthetic feature pairs (Square/Circle, and Empty/Filled), the relationship between VCR sensitivity score and measured interventional effect. Each point represents an OpenFlamingo-4B model fine-tuned on one of ten bootstrap replicates of one of five different training sets with different feature-label correlation levels.}}
  {\includegraphics[width=\linewidth]{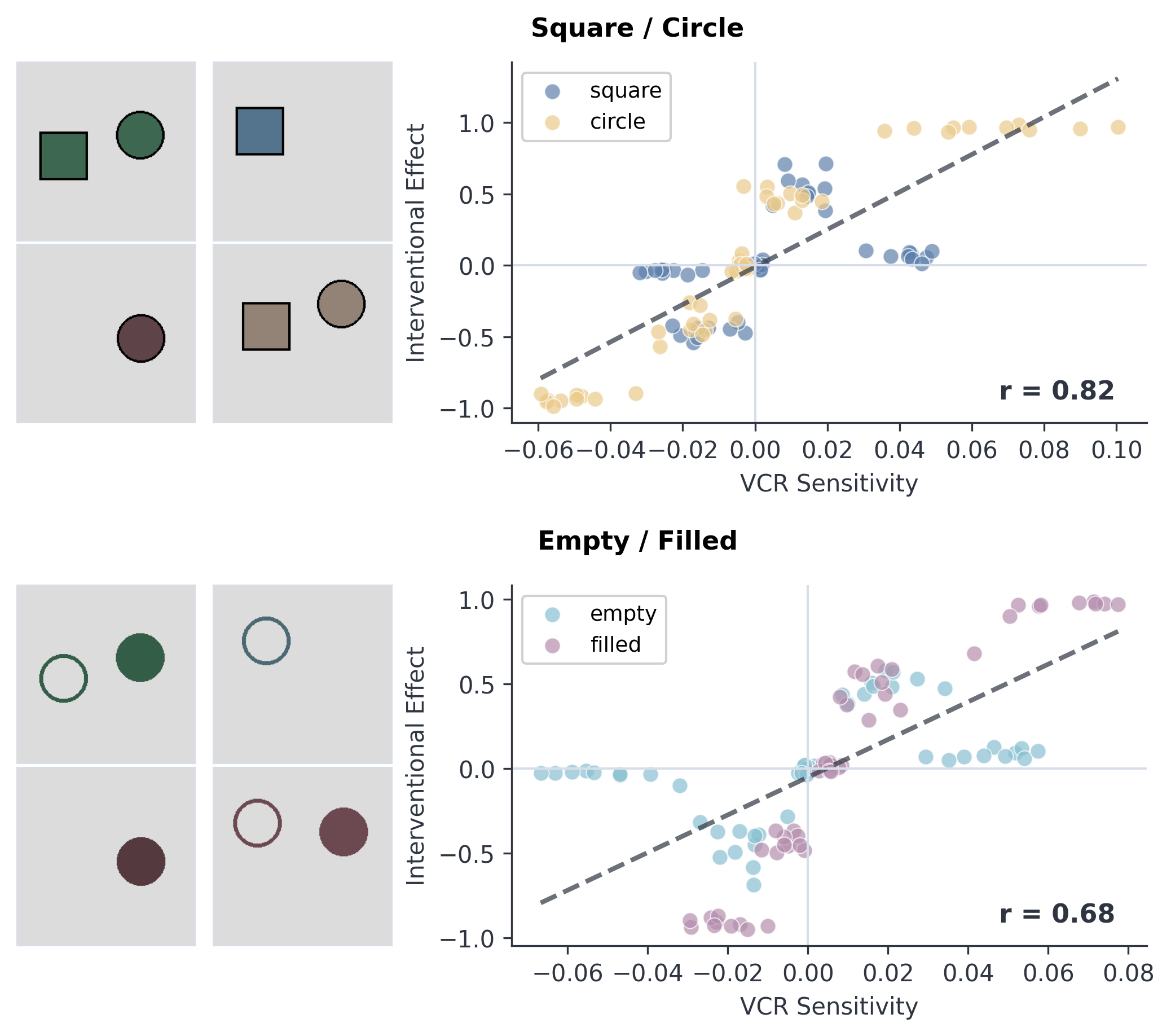}}
\end{figure}

\begin{figure}[]
\floatconts
  {fig:synthetic-data-summary}
  {\caption{Summary of VCR-intervention correlations. For each feature pair tested in an experiment with fine-tuned OpenFlamingo-4B models, except for the two pairs of feature related to position within the image, the overall correlation between VCR sensitivity and ground-truth measured interventional effect had Pearson's $r > 0.6$. Neither feature related to position within the image was significantly positively correlated.}}
  {\includegraphics[width=0.6\linewidth]{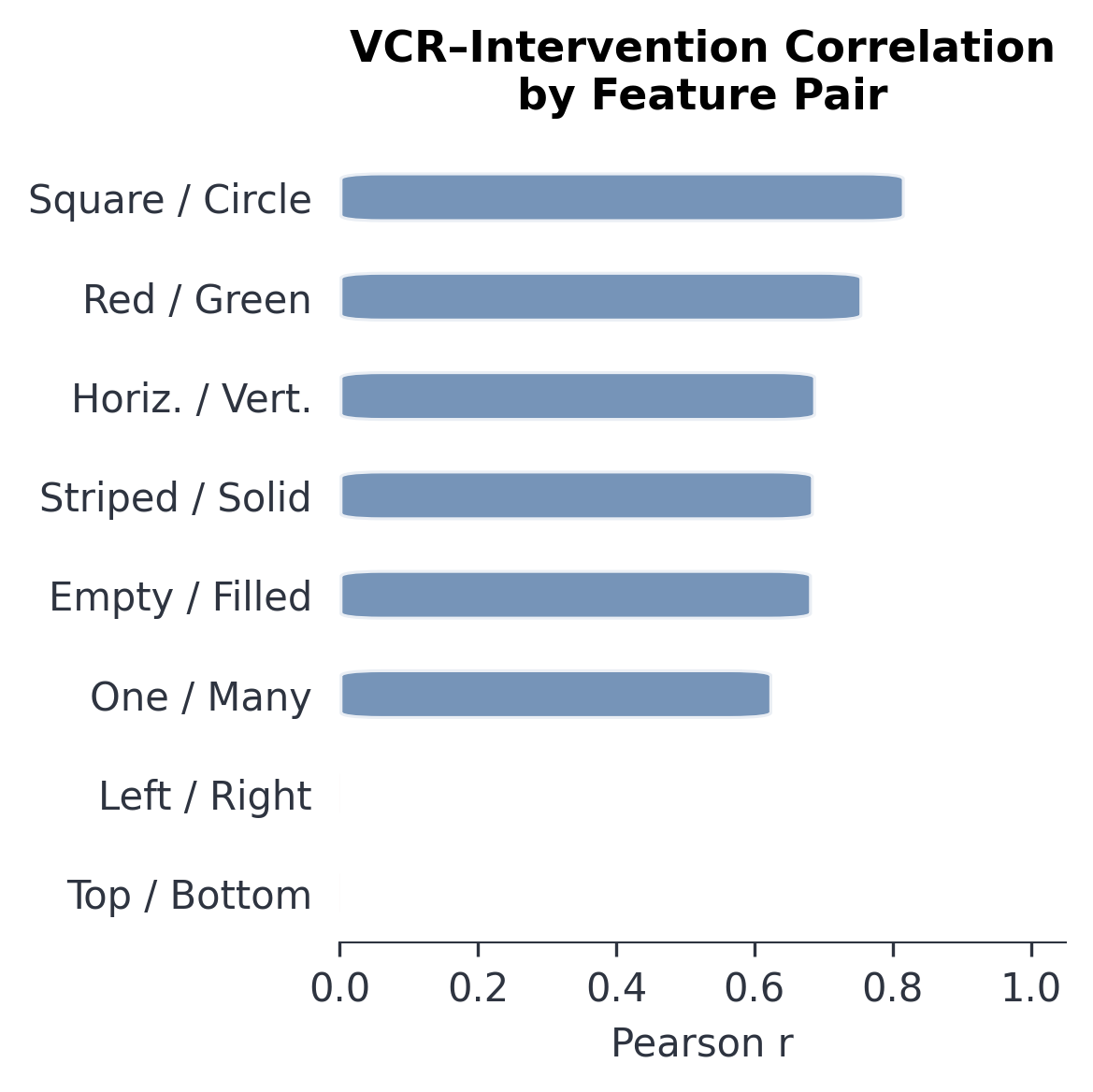}}
\end{figure}

Before applying VCR to the actual medical problems we wanted to analyze, we first needed to validate its function in a controlled setting where the ground truth is well-defined. Inspired by the Elements dataset \citep{nicolson2024explaining}, we designed a synthetic dataset experiment to allow us to control ground-truth feature-outcome relationships. We generated visual feature pairs (e.g., red vs.\ green objects, square vs.\ circle shapes) and systematically varied the correlation between each feature and a binary classification label during fine-tuning to induce different underlying model dependencies on these features (See Appendix \ref{apd:synthetic} for additional experimental details, as well as results from additional LMMs and fine-tuning settings). For each correlation condition, we fine-tuned an LMM on the synthetic data, computed ground-truth interventional effects, and compared these to VCR sensitivity scores. For each of 8 feature pairs (16 features), we tested 5 correlation levels over 10 bootstrap replicates, leading to 800 total experimental conditions. 

For an OpenFlamingo-4B model, we found a strong association between VCR sensitivity values and ground-truth interventional effects (overall Pearson's $r$: 0.53, $p: 4.9 \times 10^{-49}$). \figureref{fig:synthetic-data-main} shows example synthetic images from the training dataset, as well as detailed per-experiment results for two feature pairs, while \figureref{fig:synthetic-data-summary} shows summary results for all eight pairs of features. We repeat this set of experiments several times in Appendix \ref{apd:synthetic} with different fine-tuning settings and models (see Supplementary \figureref{fig:synthetic-of4b-last,fig:synthetic-of4b-4th,fig:synthetic-of3bi-last,fig:synthetic-of3bi-4th}), and find that the same general trend replicates -- VCR values are highly correlated with the ground truth importance of a concept. We also show that VCR and a purely correlative approach (similar to MA-MONET) \textit{both work well} in this setting where there are no confounding shortcuts or data shifts between the training data and the probe set. 

While VCR scores correlated well with actual interventional effects for 6 out of 8 feature pairs tested, synthetic data benchmarking also led us to find a failure case for our method in 2 out of 8 feature pairs (see \figureref{fig:synthetic-data-summary}). Specifically, we found that VCR struggled with concepts related to spatial locations within input images (top/bottom and left/right). This not only emphasizes the importance of rigorous benchmarking of interpretability methods, but also highlights the fact that different interpretability techniques may be complementary to each other. While the limitations of saliency maps are discussed above in the introduction, this shortcoming of our method is exactly where saliency maps excel. 

After demonstrating that VCR scores generally correlated well with interventional effects when the probe set distribution matched the training distribution, we wanted to test whether or not the method would be robust to differences in distributions. Again, this is important to be able to know if our method is identifying features that are \textit{causally important} to the model we are interpreting. For six of the eight visual feature pairs (image position features were excluded, as they did not perform well in the prior experiment without distribution shift), we designated feature A as ``reliable'' (positively correlated with the label in both training and test) and feature B as ``spurious'' (strongly negatively correlated during training with $\rho_B = -0.8$, but positively correlated at test time). A causally-meaningful interpretability method should assign positive importance to feature A and negative importance to feature B. Across 60 experimental conditions using the OpenFlamingo-4B model (6 feature pairs $\times$ 10 bootstrap replicates), VCR correctly identified the true effect of the ``spurious'' feature 92\% of the time, as compared to a correlational baseline (like MA-MONET), which only correctly identifies the true effect of the spurious feature 18\% of the time (see \figureref{fig:vcr-not-fooled}). Both approaches succeed at identifying the effect of the feature that does not undergo dataset shift between the train and probe data (98\% and 100\%). We observed a similar trend when repeating these experiments with a different LMM (OpenFlamingo-3B-Instruct) and with different fine-tuning settings (see Appendix \ref{apd:adversarial} and Supplementary \figureref{fig:adversarial-of4b-4th,fig:adversarial-of3bi-last,fig:adversarial-of3bi-4th}). This demonstrates that VCR captures \textit{causal model behavior} rather than simple input-output correlations present in the evaluation set. 

\begin{figure}[]
 % Caption and label go in the first argument and the figure contents
 % go in the second argument
\floatconts
  {fig:vcr-not-fooled}
  {\caption{VCR correctly identifies the causal effect of a spuriously correlated visual feature 92\% of the time, as compared to a CLIP-only approach that does not use model internals (akin to MA-MONET) and only identifies the correct direction 18\% of the time. Each point represents an OpenFlamingo-4B model fine-tuned on one of ten bootstrap replicates of one of six different feature-pair synthetic datasets.}}
  {\includegraphics[width=\linewidth]{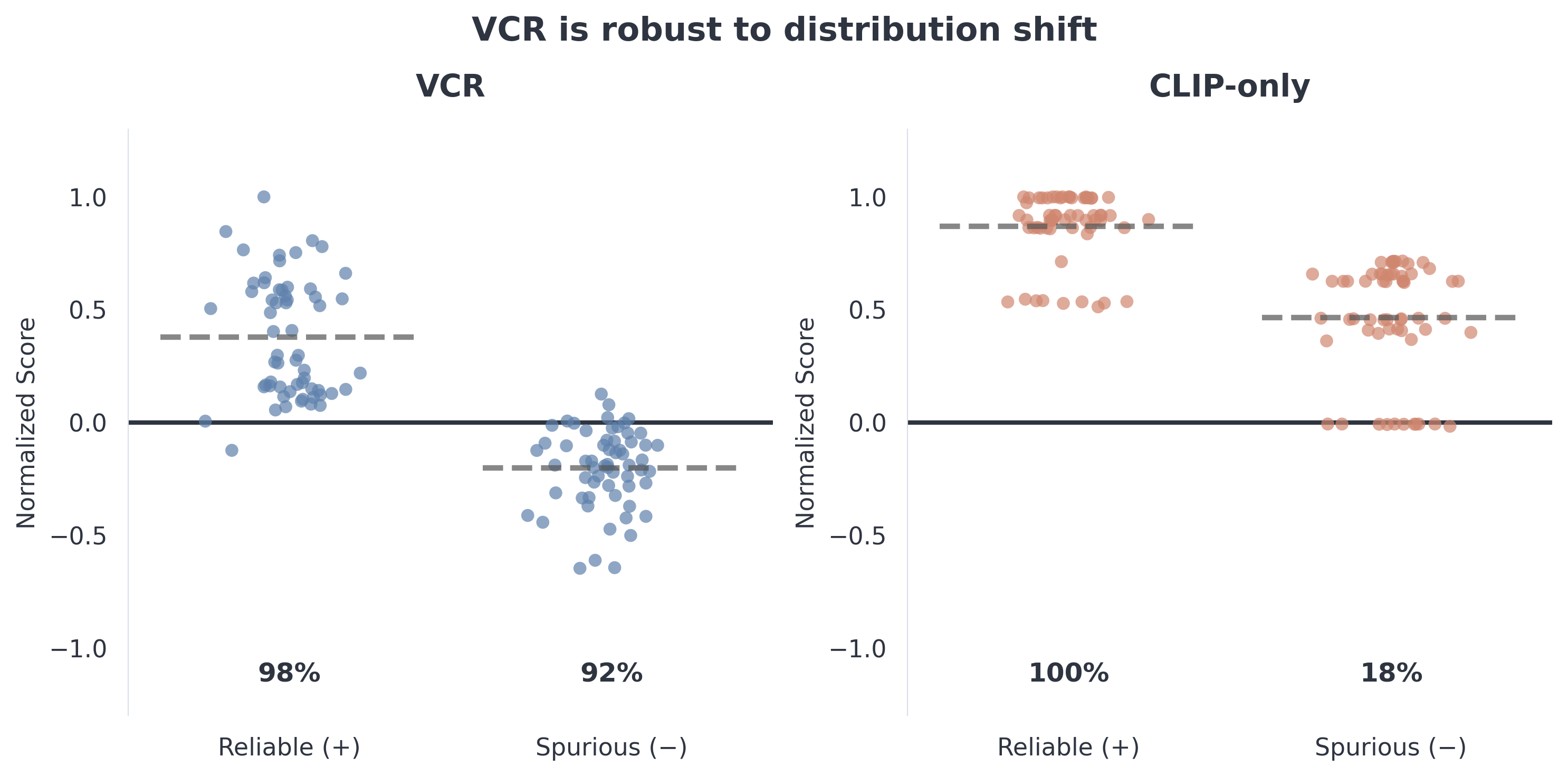}}
\end{figure}

\subsection{Understanding skin lesion classification}

After confirming the performance of our model in a synthetic data setting, we turned to applying it to the task of clinical dermatology. While numerous prior works have investigated the interpretability of dermatology image models, LMMs have provided interesting new phenomena to explain. For instance, \citet{xu2025biasicl} demonstrated a bizarre and unexpected model behavior when they prompted commercial LMMs to classify skin lesions as malignant or benign. In the zero-shot setting, ChatGPT-4o had similar predictive performance between patients with different skin type subgroups: i.e. lighter skin, Fitzpatrick Skin Type (FST) I/II, or darker skin, FST V/VI). However, when demonstrating examples were added to the prompt (In-Context Learning, \citep{radford2019language}), the model's performance improved for patients with darker skin while \textit{worsening} for patients with lighter skin.

Because our approach requires access to model gradients, it can not directly be applied to commercial models with API-only access. Therefore, we first tested to see if any open-source/open-weight LMMs recapitulated this behavior, allowing them to serve as a ``model organism'' for further investigation \cite{hubinger2023model}. We found that OpenFlamingo-3B-Instruct \citep{awadalla2023openflamingo} displayed a similar pattern to the one identified by \citet{xu2025biasicl}, where adding ICL demonstrations increased predictive performance for patients with darker skin while actually \textit{worsening} predictive performance for patients with lighter skin (\figureref{fig:auroc}).

\begin{figure}[htbp]
\floatconts
  {fig:auroc}
  {\caption{Predictive performance of OpenFlamingo-3B-Instruct for skin lesion malignancy classification across skin type subgroups as additional demonstrating examples are added in-context. Points represent the mean and shading represents one standard deviation over 3 replicates (where the specific demonstrating examples are re-sampled across replicates).}}
  {\includegraphics[width=\linewidth]{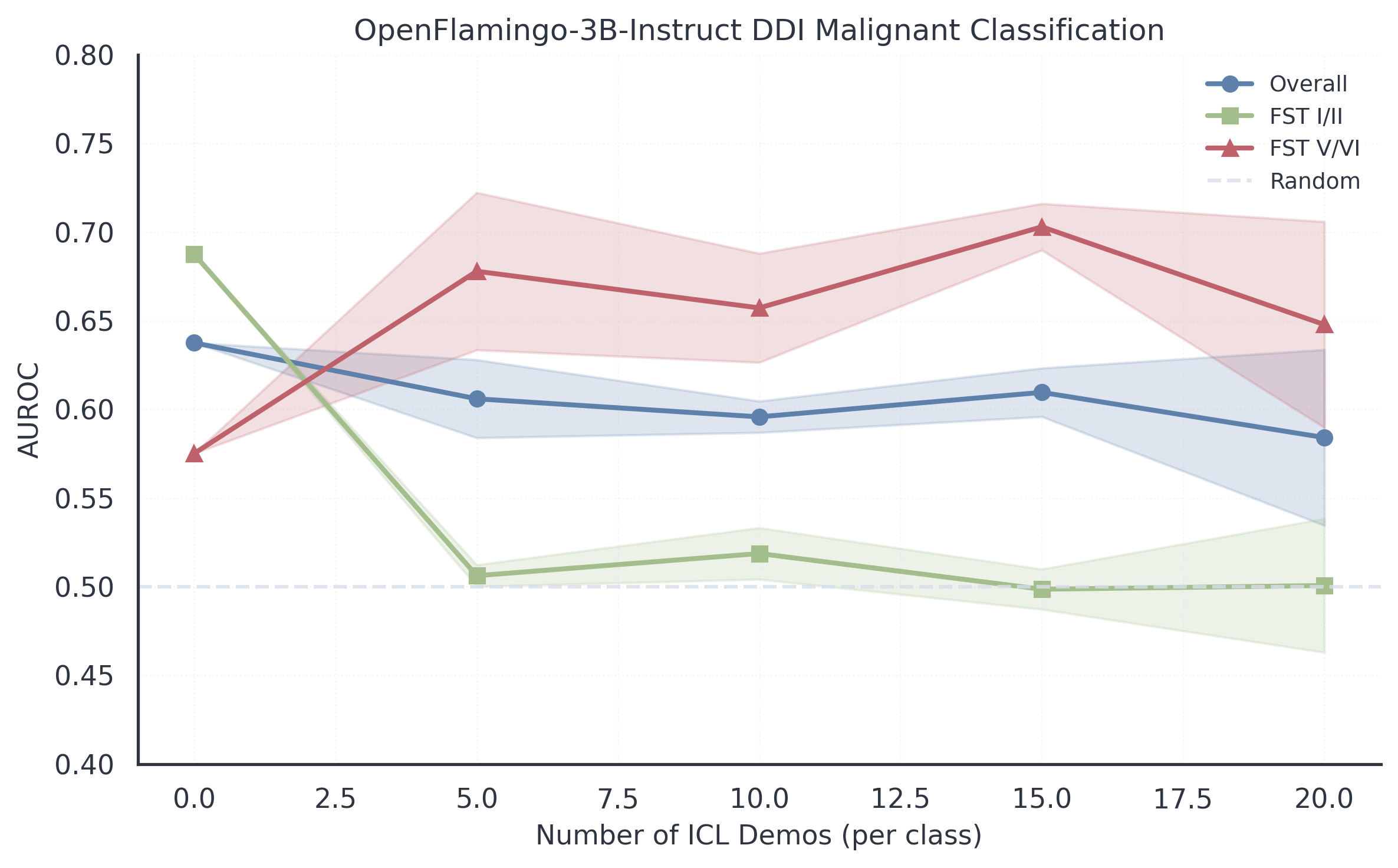}}
\end{figure}

\begin{figure}[h]
\floatconts
  {fig:top_features_of3bi}
  {\caption{Top visual features for the OpenFlamingo-3B-Instruct model in the Zero-Shot (top) and ICL (bottom) prompt settings for FST I/II (left) and FST V/VI (right) samples. Each point is one replicate of the overall VCR score on one bootstrap resampled probe set, and the bold lines indicate the mean over replicates. All features shown have significant $p$-values after Bonferroni correction.}}
  {\includegraphics[width=\linewidth]{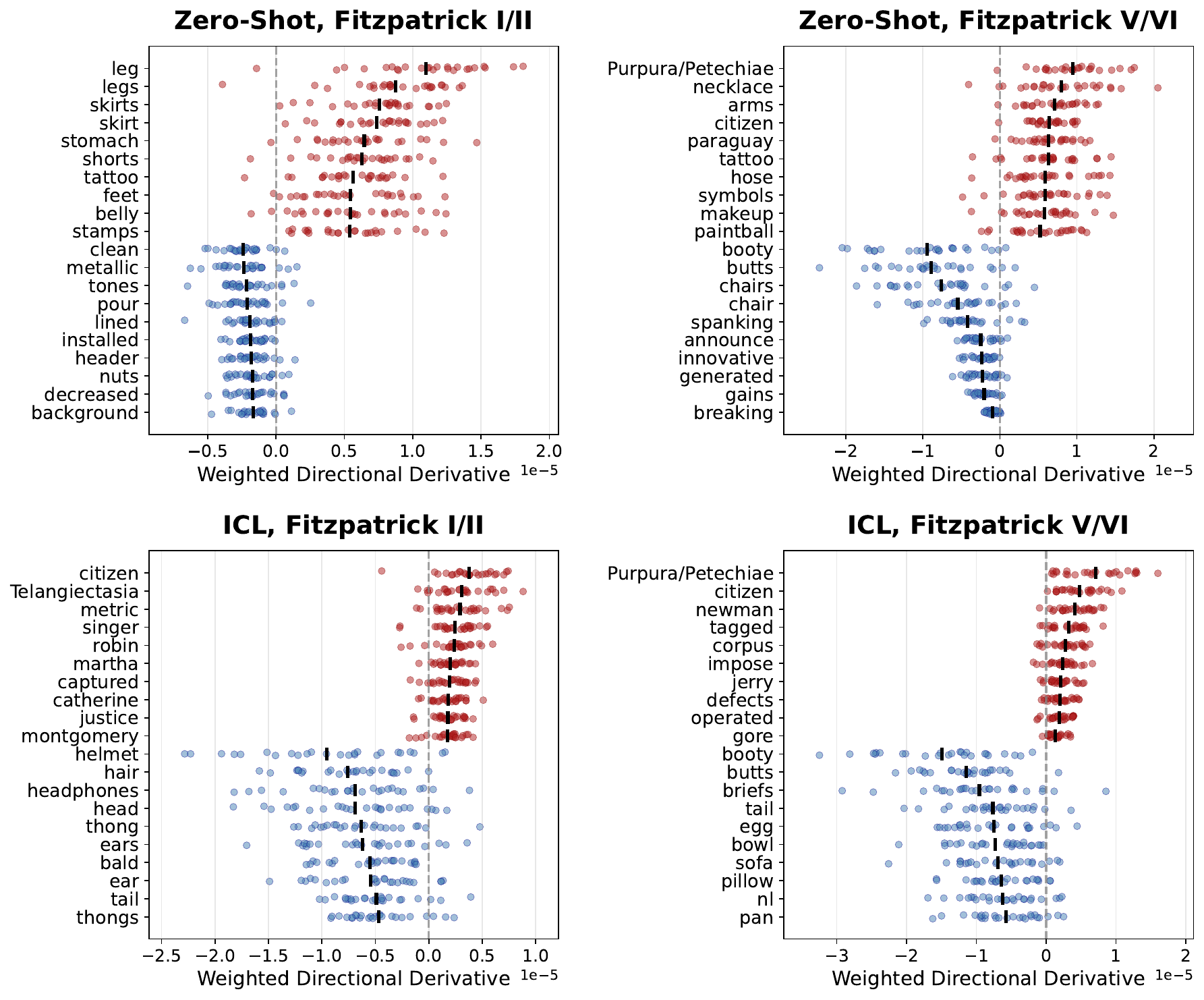}}
\end{figure}

Having confirmed that OpenFlamingo-3B-Instruct has this unexpected behavior, we next wanted to see if our VCR approach could identify differences between the visual features used by the model for FST I/II and FST V/VI skin images, which could possibly serve as an explanation. We therefore generated four sets of VCR concept explanations, one for each combination of FST I/II and FST V/VI skin images in the zero-shot and ICL settings (\figureref{fig:top_features_of3bi}). 

\begin{figure}[t]
  \centering
  \includegraphics[width=\linewidth]{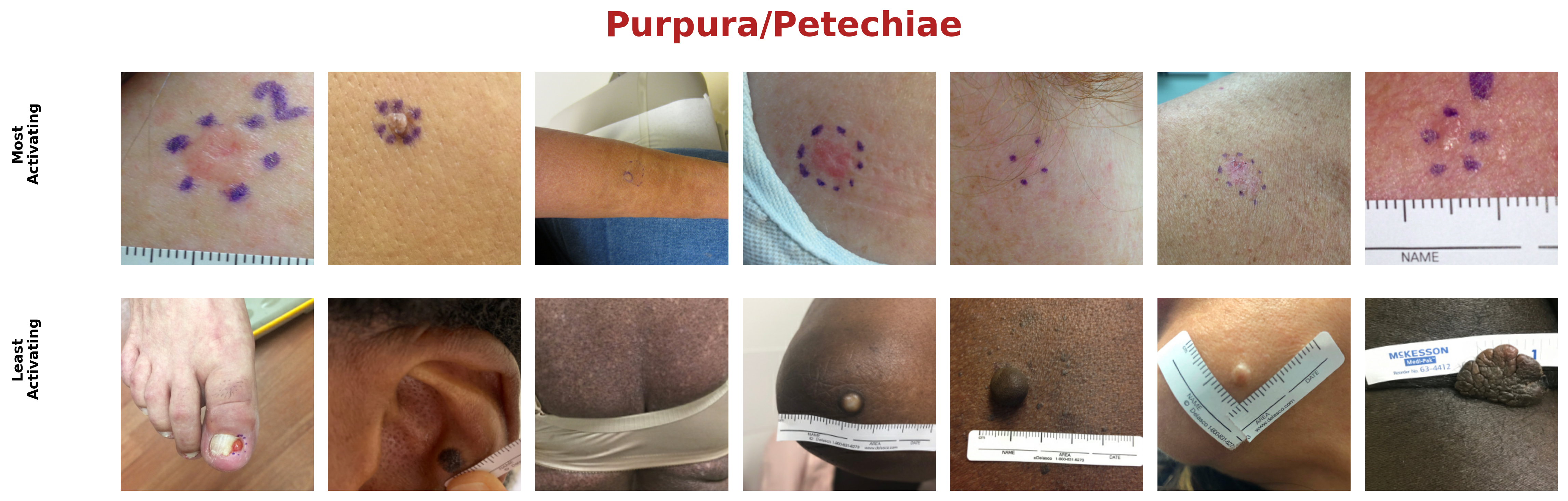}
  \caption{Most (top) and least (bottom) activating images for Purpura/Petechiae concept, which appears to be related blue/purple ink markings rather than purpura or petechiae.}
  \label{fig:purpura}
\end{figure}

\begin{figure}[t]
  \centering
  \includegraphics[width=\linewidth]{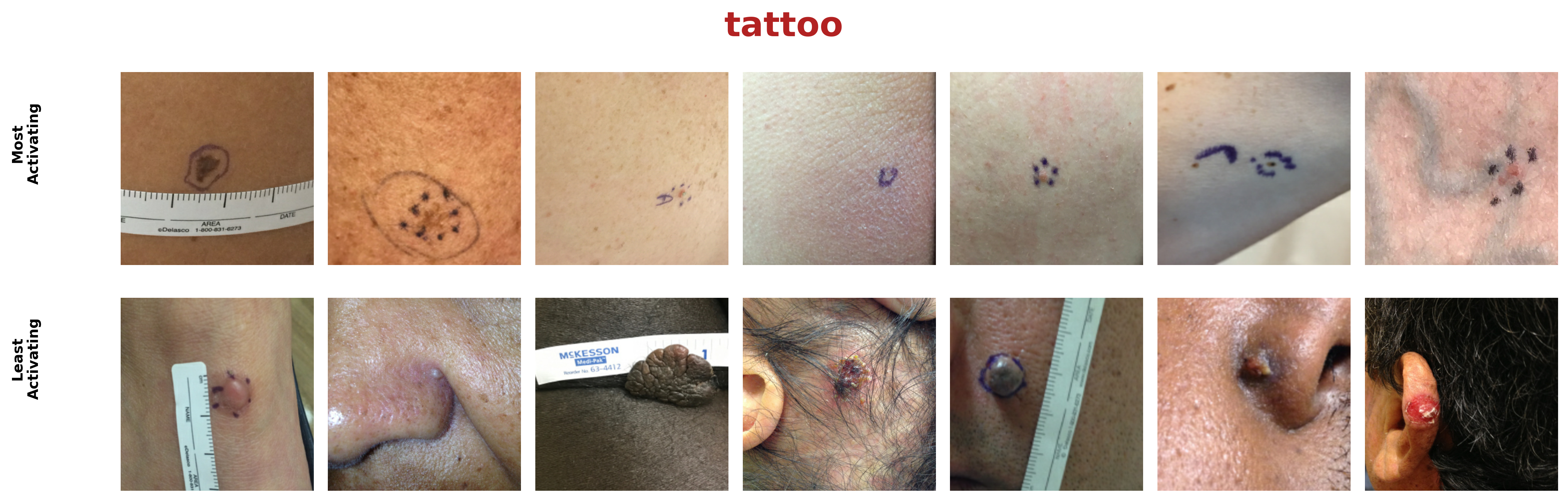}
  \caption{Most (top) and least (bottom) activating images for tattoo concept, which also appears to be related to blue/purple ink markings.}
  \label{fig:tattoo}
\end{figure}

Within the set of top concepts increasing the probability of malignancy, we saw that for all settings except ICL + Fitzpatrick I/II, the concepts ``tattoo'' and/or ``Purpura/Petechiae'' were statistically significant. After plotting the most and least activating images for both of these concept vectors, we saw that both concepts corresponded to blue/purple skin markings \figureref{fig:tattoo}, which are frequently used by dermatologists to mark skin lesions for biopsy, and hence serve as a shortcut for malignancy prediction. This led us to a testable hypothesis: adding blue/purple skin markings should increase the predicted probability of ``malignant'' for the OpenFlamingo-3B-Instruct model for FST V/VI patient samples in both the zero-shot and ICL setting, but should only do so for FST I/II patient samples in the zero-shot setting.

\begin{figure}[t]
  \centering
  \includegraphics[width=\linewidth]{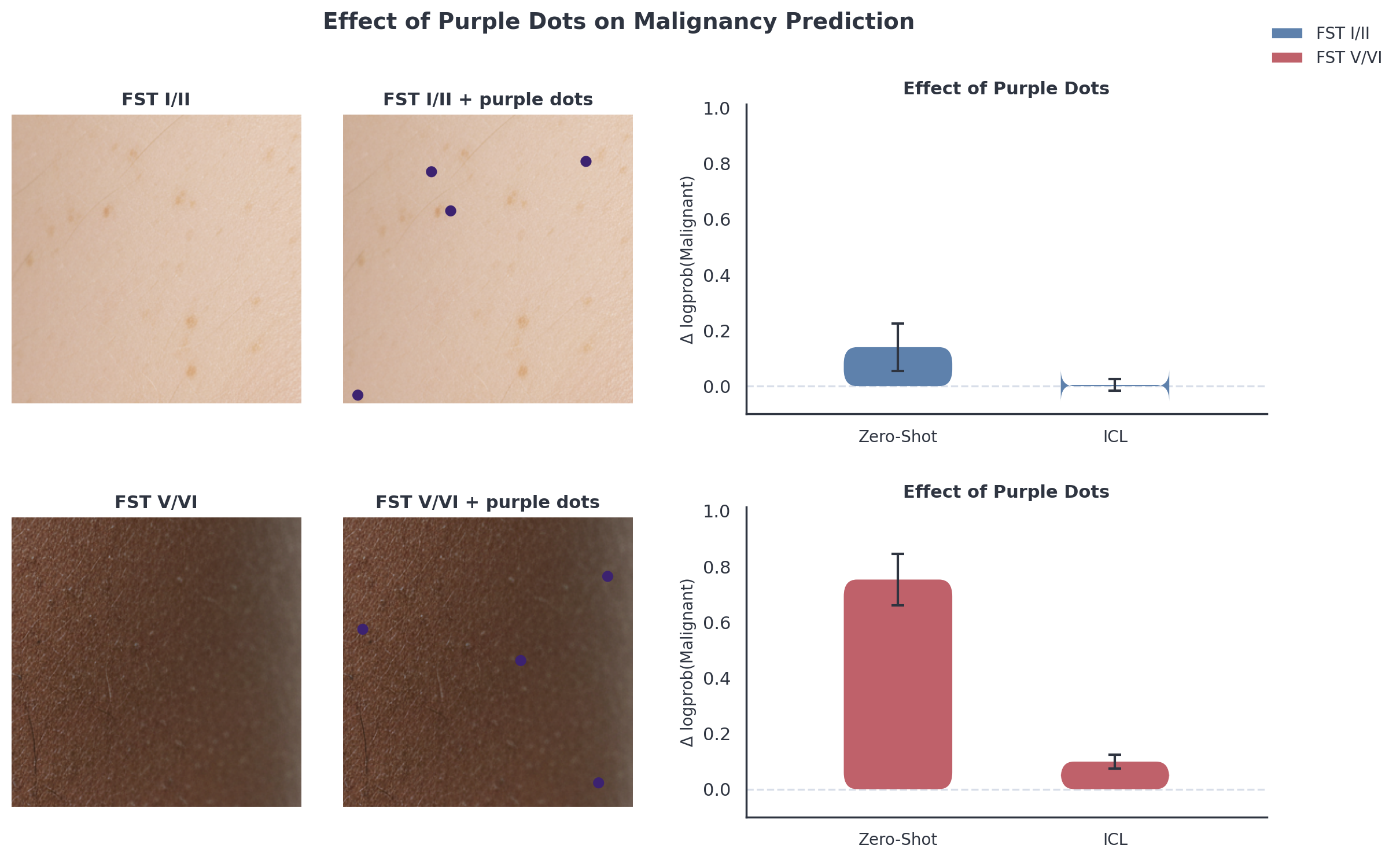}
  \caption{(Left) Examples of skin images before and after intervening by adding purple/blue dots. (Right) As predicted by the VCR explanations, the purple dots have an impact in both the zero-shot and ICL settings for FST V/VI patients, but only in the zero-shot setting for FST I/II patients.}
  \label{fig:purple_dot_intervention}
\end{figure}

We confirmed this hypothesis by manually adding these blue/purple skin markings to neutral background skin images. We generated the background skin images for each skin type using Gemini-3 (see Appendix for further details), and added the skin markings using Python Imaging Library (PIL). We find that this manual intervention confirms the exact hypothesis generated by our VCR concepts: the blue/purple skin markings are persistently important for FST V/VI after demos are added to the prompt in the ICL setting, but lose their saliency for FST I/II samples.

\begin{figure}[t]
  \centering
  \includegraphics[width=0.6\linewidth]{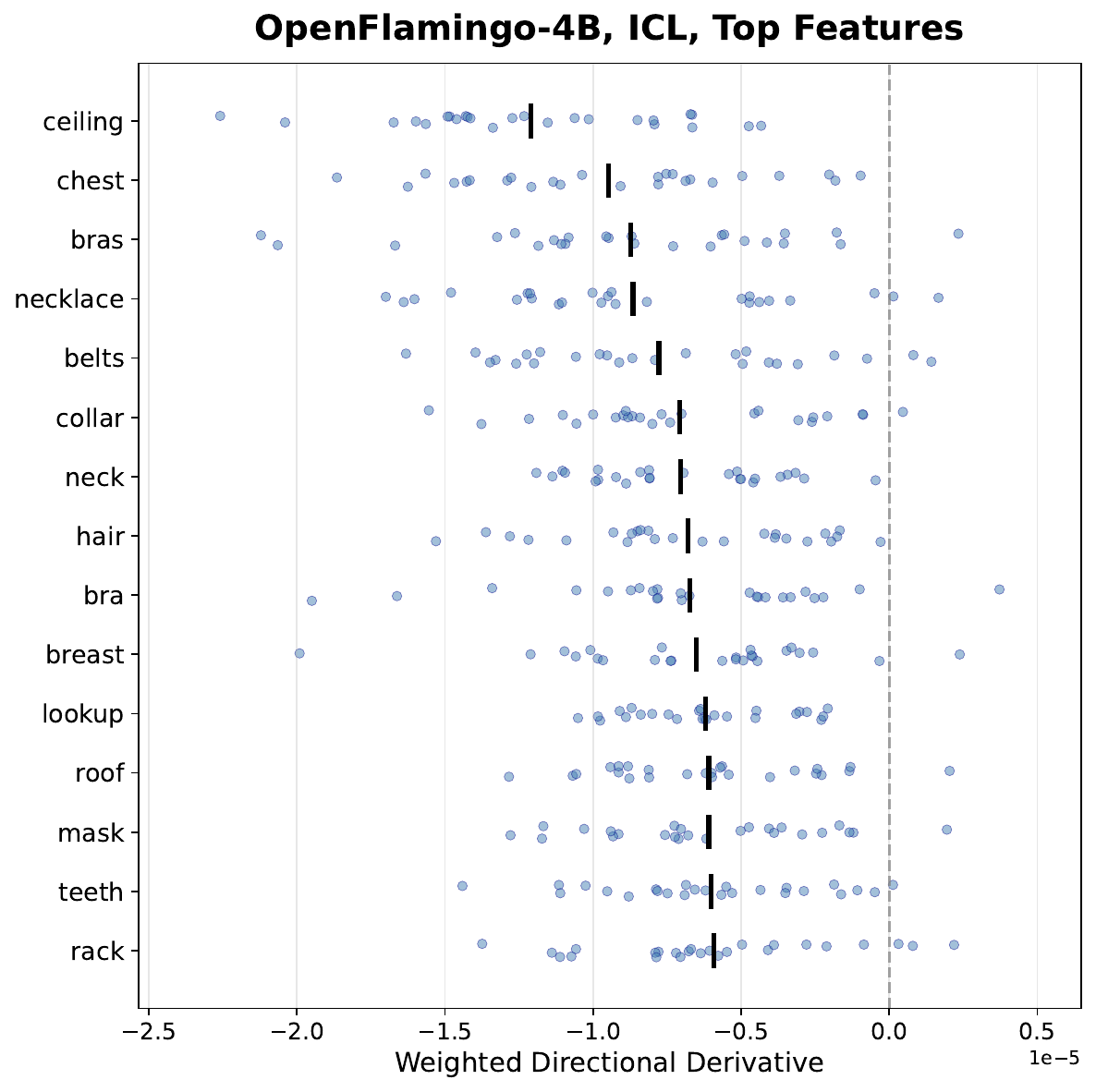}
  \caption{For the OpenFlamingo-4B model prompted with an ICL prompt, the top statistically significant visual features decreasing the probability of malignancy.}
  \label{fig:OF4B_bulk_exp}
\end{figure}

In addition to generating hypotheses about differences between performance for different demographic subgroups, we also wanted to see what features these models might rely upon more generally across all samples. For example, by plotting the top concepts decreasing the probability of malignancy for OpenFlamingo-4B, we saw many examples that seemed to be related to the background body part upon which the lesion was found. This was of particular interest to us as this is the sort of feature that prior methods would have struggled to identify; saliency maps would struggle to localize a feature spread so diffusely throughout an image, and counterfactual image generation approaches would struggle to maintain cycle-consistency while changing so many pixels within an image.

\begin{figure}[t]
  \centering
  \includegraphics[width=\linewidth]{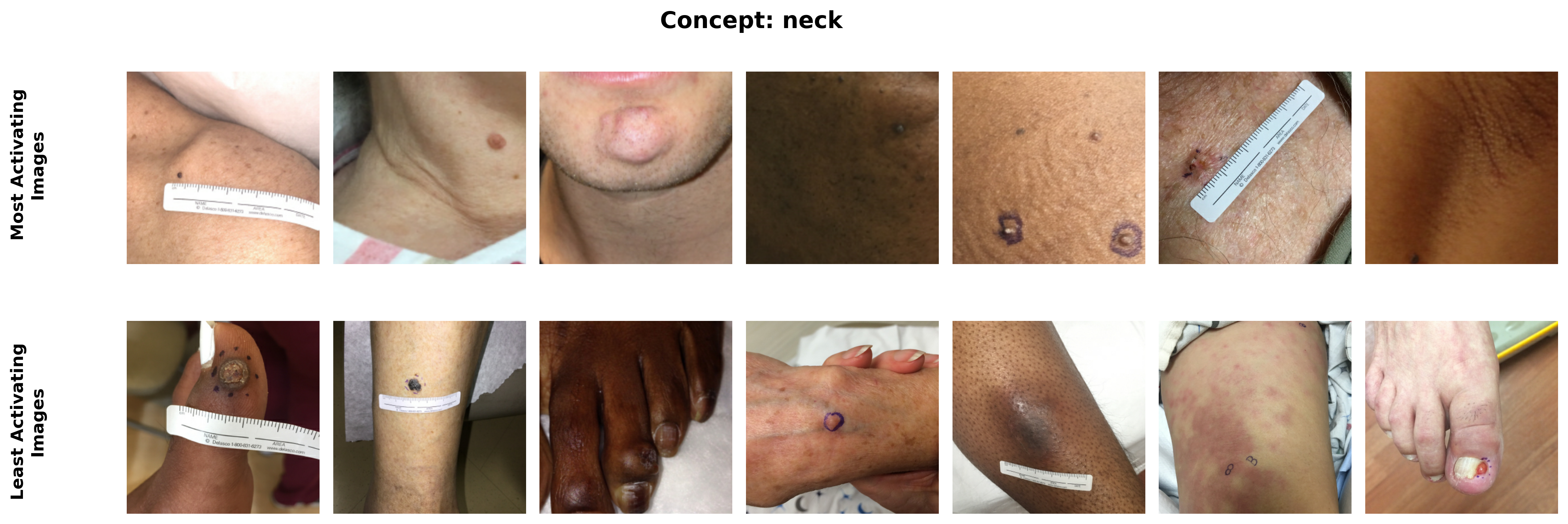}
  \caption{Most (top) and least (bottom) activating images for neck concept.}
  \label{fig:neck}
\end{figure}

\begin{figure}[t]
  \centering
  \includegraphics[width=\linewidth]{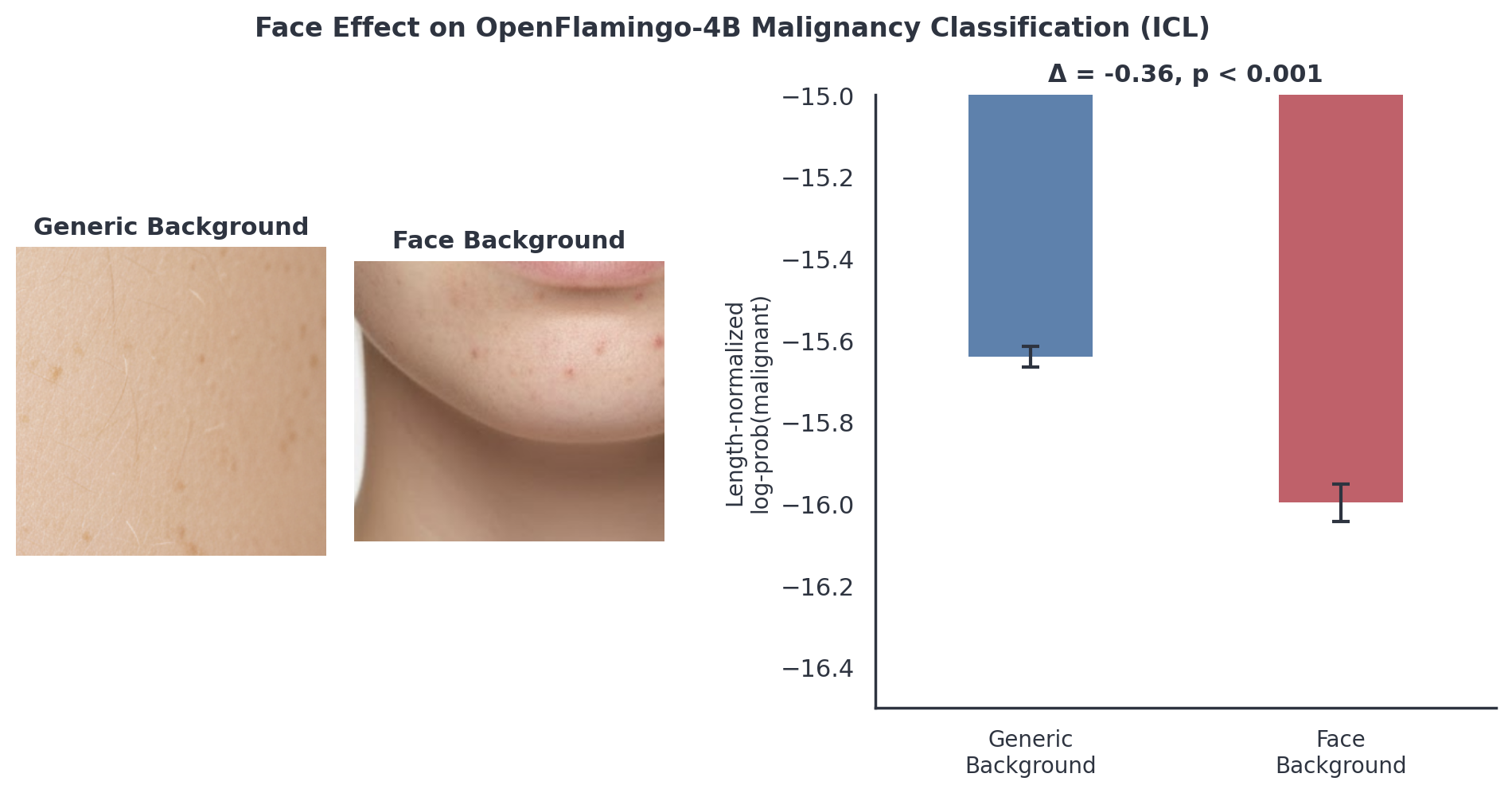}
  \caption{OpenFlamingo-4B is more likely to predict a lesion on a neutral/generic skin background is malignant than a lesion on the background of a patient's face/neck.}
  \label{fig:face_exp}
\end{figure}

Among the top concepts in \figureref{fig:OF4B_bulk_exp}, we found multiple visual features that seemed to be associated with lesions found on the face (e.g. ``collar,'' ``neck,'' and ``teeth''), and the top activating examples for these concepts confirmed that these concepts were activated by images containing patients' faces/necks (see \figureref{fig:neck}). This led us to the hypothesis that we should expect OpenFlamingo-4B to output on average a lower probability of malignancy for images with face backgrounds than generic skin backgrounds. We tested this by using the same generic skin background images generated by Gemini-3 in the prior experiment, and comparing the average model output on these images to another set of generated skin images of patients' faces/necks. We found that the probability assigned to the ``malignant'' completion was significantly lower for the face background images than the generic skin background images (see \figureref{fig:face_exp}).

\begin{figure}[t]
  \centering
  \includegraphics[width=\linewidth]{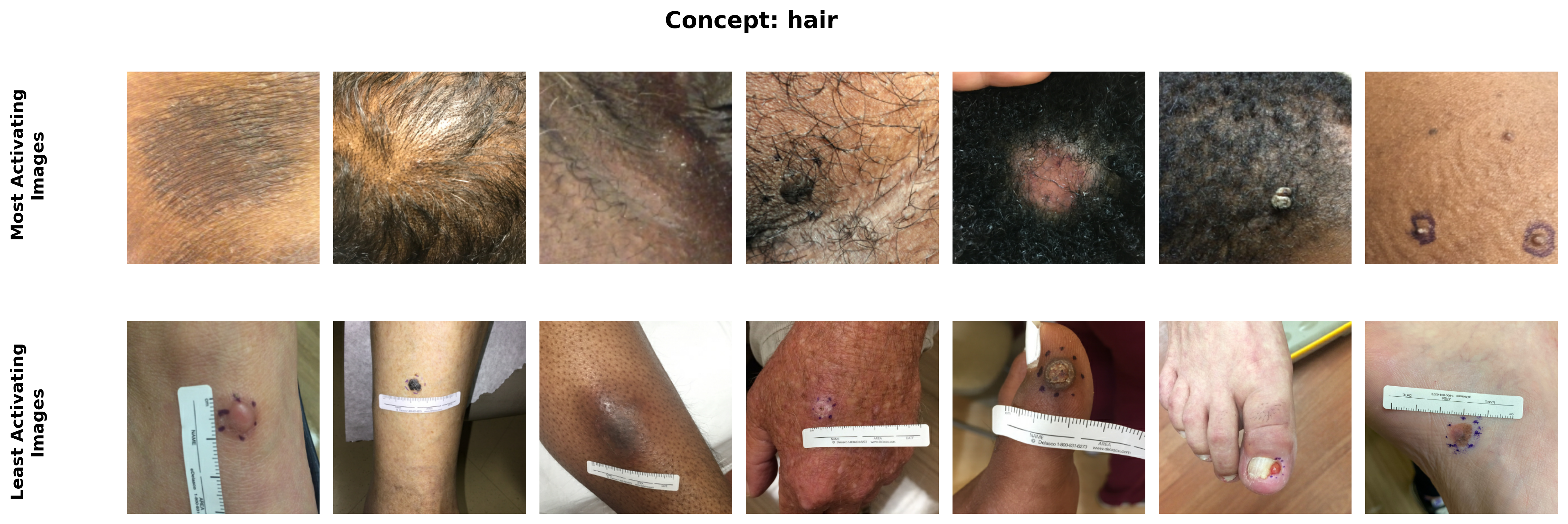}
  \caption{Most (top) and least (bottom) activating images for hair concept.}
  \label{fig:hair}
\end{figure}

\begin{figure}[t]
  \centering
  \includegraphics[width=\linewidth]{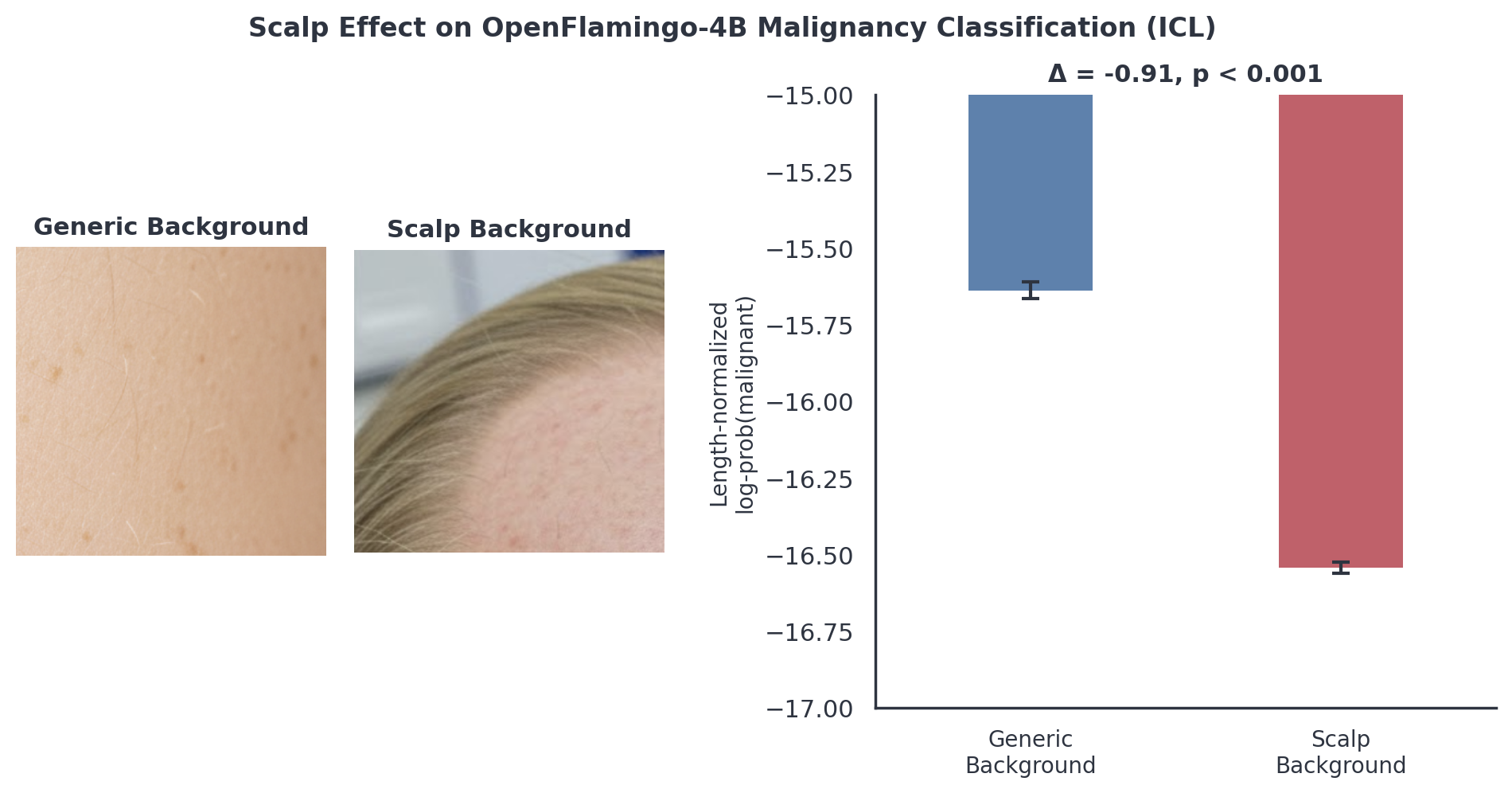}
  \caption{OpenFlamingo-4B is more likely to predict a lesion on a neutral/generic skin background is malignant than a lesion on the background of a patient's scalp.}
  \label{fig:scalp_exp}
\end{figure}

Another feature we investigated was a concept labeled by our approach as ``hair.'' When we examined the top activating images for this concept, however, it appeared that the concept activated more specifically for images of scalps than ``hair'' alone (see \figureref{fig:hair}). This led us to another testable hypothesis: that we should expect OpenFlamingo-4B to output a lower probability of malignancy on average for images with scalp backgrounds than generic skin backgrounds. When using the same testing protocol described above and in Appendix \ref{apd:scalp}, we found that the probability assigned to the ``malignant'' completion was significantly lower for the scalp background images than the generic skin background images (see \figureref{fig:scalp_exp}).

\subsection{Additional supplemental experiments}

In addition to our detailed experiments on a synthetic data benchmark and a dermatology dataset, we included several supplemental experiments to showcase the versatility of our method. 

For instance, we applied our method to the CheXpert dataset of chest radiographs \citep{rajpurkar2017chexnet}. We prompted the model to differentiate normal from abnormal chest radiographs. In response to this prompt, we found that one of the top concepts used by the model was named ``wiring.'' When we examine the top activating examples for this concept, we see that this concepts activates for radiographs with medical support devices such as pacemaker wires and EKG leads (\figureref{fig:wiring}). This is an obvious shortcut, in the sense that the models are again looking at visual clues that a doctor was suspicious that the patient was sick rather than looking for anatomic/pathologic evidence that the patient is sick. 

To demonstrate the broad applicability of our method outside of medical tasks, we also experiment with applying our method in the context of non-medical prompts using natural images from the publicly available Imagenette dataset (\figureref{fig:logos,fig:dotted}). Finally, to demonstrate the computational efficiency of our approach, we included several experiments testing how the computational cost of our method scales with concepts in the concept set (\figureref{fig:concept_scaling}) and images in the probe set (\figureref{fig:probe_scaling}).

\section{Discussion}

In this work, we introduced Visual Concept Ranking (VCR), a method for identifying the visual concepts that large multimodal models depend upon when completing specific tasks. We validated VCR on synthetic datasets where ground-truth feature dependencies were known, demonstrating strong correlations between VCR sensitivity scores and actual interventional effects. Critically, we showed that VCR correctly identifies causal model dependencies even under distribution shift between training and evaluation data, a setting where purely correlational approaches fail. We then applied VCR to investigate LMM behavior on clinical dermatology images, generating hypotheses about model shortcuts that we subsequently confirmed through manual image interventions.

One of our central findings relates to the unexpected demographic disparities that emerge when LMMs are prompted with in-context learning demonstrations. VCR analysis revealed that concepts related to blue/purple skin markings, which serve as shortcuts for malignancy prediction since these markings are often used by dermatologists to identify lesions for biopsy, remained significant predictors for FST V/VI patients across both zero-shot and ICL settings but lost significance for FST I/II patients in the ICL setting. While the reliance of dermatology AI models on ink markings has been documented previously \citep{bissoto2020debiasing}, our findings demonstrate that this dependency is not static: it varies with the specific prompt used (zero-shot versus ICL) and is not homogenous across demographic groups. This suggests that the addition of demonstration examples may alter the model's reliance on shortcut features differently across skin type subgroups, perhaps because demonstration images contain different distributions of these shortcuts across skin types. This has significant implications for clinical deployment, as careful curation of demonstration examples may be necessary to ensure equitable model performance.

Beyond demographic disparities, our analysis uncovered model dependencies on anatomical location that could compromise clinical utility. VCR identified concepts such as "neck" and "hair" as significantly decreasing predicted malignancy probability, and interventional experiments confirmed that lesions presented on face or scalp backgrounds received lower malignancy predictions than identical lesions on generic skin backgrounds. If models systematically underestimate malignancy risk for lesions on certain body sites, this could lead to missed diagnoses. These location-based biases represent the kind of shortcut behavior that would be difficult to detect through standard performance benchmarking alone.

A key strength of VCR relative to prior approaches is its robustness to distribution shift. Methods that rely on correlations between input features and model outputs will report features as important if they correlate with outcomes in the evaluation set, regardless of whether the model actually uses those features. VCR directly measures model sensitivity through gradients and activation patterns, enabling it to distinguish between features that are merely correlated with outcomes and features that causally influence predictions. VCR also offers complementary strengths to other interpretability approaches: saliency maps excel at spatially localized features but struggle with diffuse concepts like background body part, while counterfactual image generation can test specific hypotheses but struggles to maintain cycle-consistency when modifying large image regions.

Several limitations warrant discussion. VCR requires access to model gradients, precluding direct application to API-only commercial models. Our synthetic benchmarks revealed that VCR struggles with concepts related to spatial position within images, a setting where traditional saliency methods remain preferable. 

The semantic labels assigned to concept vectors may not perfectly capture their true meaning, as we observed with the ``Purpura/Petechiae'' and ``tattoo'' labels activating most strongly for dermatologic skin ink markings, and a ``hair" concept activating most strongly for scalp images rather than hair per se. This underscores that VCR is best understood as a hypothesis-generation tool requiring human interpretation and validation rather than a fully automated auditing system.

Several avenues for future work emerge from this study, including combining VCR with activation steering methods to enable correction of undesirable behaviors and developing improved methods for encoding spatial concepts. As LMMs become increasingly prevalent in safety-critical domains like healthcare, methods that illuminate their internal mechanisms will be essential for building justified trust in their predictions.

\bibliography{chil-sample}

\appendix

\section{Visual Summary of VCR Algorithm}\label{apd:vcr-algo}

In addition to the textual description of the VCR method in the main text, we also provide a visual summary of the main steps in the algorithm in \figureref{fig:visual-algo}.

\begin{figure*}[h]
\floatconts
  {fig:visual-algo}
  {\caption{Visual summary of our method, VCR, to automatically interpret large multimodal models (LMMs) such as OpenFlamingo (center, grey box). \textbf{1}, A separate, pretrained vision-language model (e.g. OpenCLIP) is used to label a probe set of $N$ images with continuous values reflecting how much a set of $K$ concepts are represented in each image, generating a concept label matrix $Y \in \mathbb{R}^{N \times K}$. \textbf{2}, For each image in the probe set, the corresponding activations from the LMM are extracted into an activation matrix $A \in \mathbb{R}^{N \times D}$, where $D$ is the residual stream dimension. \textbf{3}, Concept activation vectors for each concept are calculated by learning linear models to predict each concept label in $Y$ from the activations in $A$. \textbf{4}, To calculate the importance of these concept vectors, the directional derivative of the length-normalized log-probs of the target completion of the LMM are taken with respect to each concept, then averaged over all images in the probe set. This entire process is then repeated a number of times with bootstrap resampled versions of the probe set to allow for statistical significance testing, ensuring that the concept importance scores have the same direction over the bootstrap resampled probe sets.}}
  {\includegraphics[width=\textwidth]{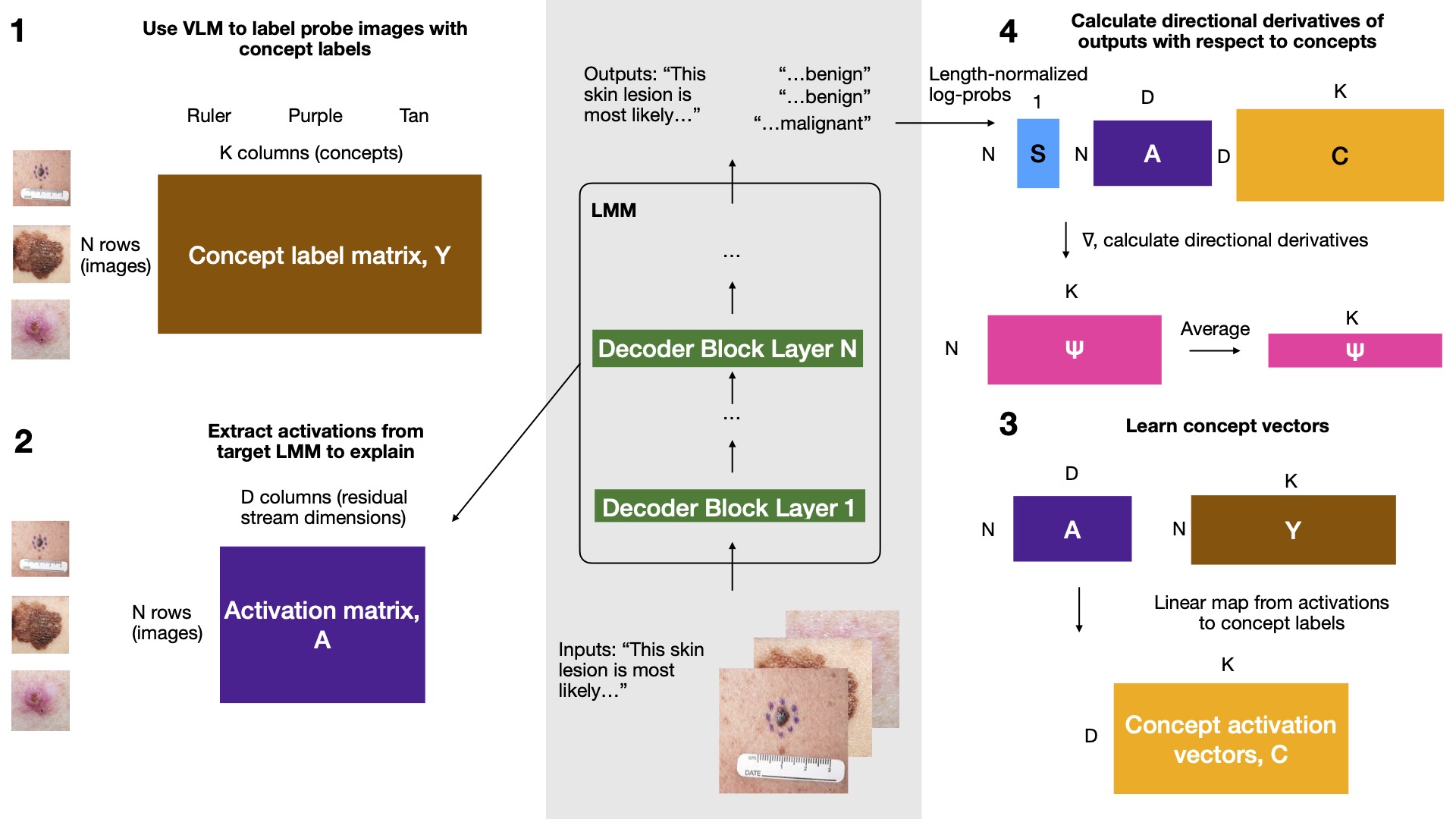}}
\end{figure*}

\section{Detailed Methods for Synthetic VCR Validation}
\label{apd:synthetic}
%%%%%%%%%%%%%%%%%%%%%%%%%%%%%%%%%%%%%%%%%%%%%%%%%%%%%%%%%%%%%%%%%%%%%%%%%%%%%%%

\subsection{Overview and Motivation}

The core challenge in validating interpretability methods is establishing ground truth: in natural datasets, we rarely know which visual features truly cause a model's predictions versus which are merely correlated with outputs. To address this, we constructed synthetic datasets where we precisely control feature-label correlations, enabling direct comparison between VCR's predicted concept importance and measured causal effects.

\subsection{Visual Feature Pairs}

We designed eight binary feature pairs:
\begin{enumerate}
    \item \textbf{Red vs.\ green}: Colored rectangles (RGB 220,50,50 vs.\ 50,180,50) on neutral gray background, with 2-pixel dark outline and small elliptical distractors.
    \item \textbf{Left vs.\ right}: Single ellipse positioned in left half ($x \in [10, W/2 - 60]$) vs.\ right half ($x \in [W/2 + 10, W - 60]$) of the image.
    \item \textbf{One vs.\ many}: Single large ellipse (50px diameter) vs.\ cluster of 6--10 small ellipses (18--20px diameter).
    \item \textbf{Horizontal vs.\ vertical}: Rectangular bar oriented horizontally (width $\gg$ height) vs.\ vertically (height $\gg$ width).
    \item \textbf{Square vs.\ circle}: 60px square vs.\ 60px diameter circle with 3-pixel black outline.
    \item \textbf{Empty vs.\ filled}: Hollow ellipse (5px outline, no fill) vs.\ solid filled ellipse. VCR concept names: ``loops'' (empty), ``spots'' (filled).
    \item \textbf{Striped vs.\ solid}: Rectangle with vertical stripes (3px width, 8px spacing) vs.\ uniform solid fill. VCR concept names: ``bars'' (striped), ``cube'' (solid).
    \item \textbf{Top vs.\ bottom}: Ellipse positioned in upper half ($y \in [15, H/2 - 65]$) vs.\ lower half ($y \in [H/2 + 15, H - 65]$).
\end{enumerate}

While we initially computed VCR scores using the concept names we expected, we found that for two experiments (``empty'' vs ``filled'', and ``striped'' vs ``solid''), the anticipated concept names were not significantly important. The cause for this was immediately resolved when examining the top significant features. For the Empty vs filled experiment, we saw that the top features were ``loops'' and ``spots,'' rather than the names we anticipated these features would have. Likewise, for Striped and solid the top features were ``bars'' and ''cube,'' which makes sense given that all of the images in our synthetic dataset for that example were rectangular. 

\subsection{Image Generation}

All images were generated at $224 \times 224$ pixels using PIL/Pillow. Each generator function accepts boolean flags (\texttt{has\_a}, \texttt{has\_b}) indicating presence of each feature, along with a random seed for reproducibility. Images include controlled randomization:
\begin{itemize}
    \item Object positions sampled uniformly within designated regions
    \item Object colors sampled from RGB range $[50, 150]^3$ (except for red/green pair)
    \item Background colors set to neutral grays (RGB values 200--235)
    \item Small distractor shapes added where appropriate
\end{itemize}

\subsection{Dataset Construction}

For each feature pair, we created training sets with systematically varied feature-label correlations. The correlation parameter $\rho_A \in \{-1.0, -0.5, 0.0, +0.5, +1.0\}$ controls feature A's correlation with the positive label, while $\rho_B = -\rho_A$ ensures anticorrelated features.

Training sets contained $N_{\text{train}} = 400$ images (200 positive, 200 negative). For positive-labeled images, the number of images containing feature A was:
\begin{equation}
    n_{A|\text{pos}} = \left\lfloor \frac{N_{\text{pos}}}{2} \cdot (\rho_A + 1) \right\rfloor
\end{equation}
For negative-labeled images:
\begin{equation}
    n_{A|\text{neg}} = \left\lfloor \frac{N_{\text{neg}}}{2} \cdot (1 - (\rho_A + 1)/2) \right\rfloor
\end{equation}

Test sets contained 200 images with perfectly balanced feature combinations: 50 images for each of the four $(\texttt{has\_a}, \texttt{has\_b}) \in \{\text{True}, \text{False}\}^2$ conditions, enabling unbiased estimation of interventional effects. Test images used seeds offset by 10,000 from training images to ensure no overlap.

\subsection{Model Architecture and Fine-tuning}

We used both OpenFlamingo-3B-Instruct and OpenFlamingo-4B as the base large multimodal models for our experiments, accessed via our custom \texttt{FlamingoAPI} wrapper. The model architecture consists of a vision encoder and language model with cross-attention layers for multimodal fusion.

\paragraph{Fine-tuning configuration:}
\begin{itemize}
    \item \textbf{Trainable parameters}: We ran experiments with either the final transformer block (\texttt{model.lang\_encoder.transformer.blocks}) unfrozen, or with the fourth from final transformer block unfrozen; all other parameters remained frozen.
    \item \textbf{Optimizer}: AdamW with learning rate $\eta = 10^{-4}$
    \item \textbf{Batch size}: 8
    \item \textbf{Epochs}: 5
    \item \textbf{Loss function}: Cross-entropy on next-token prediction
    \item \textbf{Training prompt}: \texttt{"<image>This image is \{label\}"} where $\text{label} \in \{\text{positive}, \text{negative}\}$
\end{itemize}

To ensure robust estimates, we performed 10 bootstrap replicates per condition, resampling training data with replacement (using \texttt{pandas.DataFrame.sample} with \texttt{replace=True}) while keeping test sets fixed.

\subsection{Interventional Effect Measurement}

Ground-truth interventional effects were computed on the balanced test set after fine-tuning. For each test image $i$:

\begin{enumerate}
    \item Extract next-token logits given prompt \texttt{"<image>This image is"}
    \item Compute normalized probability:
    \begin{equation}
        P_i(\text{positive}) = \frac{\exp(\ell_{\text{pos}})}{\exp(\ell_{\text{pos}}) + \exp(\ell_{\text{neg}})}
    \end{equation}
    where $\ell_{\text{pos}}$ and $\ell_{\text{neg}}$ are logits for the tokens `` positive'' and `` negative'' (with leading space).
\end{enumerate}

The interventional effect for feature $X \in \{A, B\}$ was calculated as:
\begin{equation}
    \mathbb{E}[P(\text{pos}) \mid X = \text{present}] - \mathbb{E}[P(\text{pos}) \mid X = \text{absent}]
\end{equation}
where expectations are computed over the balanced test set. Positive effects indicate the feature increases $P(\text{positive})$; negative effects indicate suppression.

\subsection{VCR Sensitivity Computation}

VCR analysis was performed using the \texttt{ConceptAnalyzer} class with the following configuration:

\paragraph{Concept embedding:}
\begin{itemize}
    \item Concept vocabulary: Google 10,000 English words \citep{google10000english}
    \item Embedding model: OpenCLIP (via \texttt{CLIPEmbedder} utility class)
    \item Image-concept similarity: Inner product between OpenCLIP image embeddings and text embeddings
\end{itemize}

\paragraph{Activation collection:}
\begin{itemize}
    \item Hook layer: Always the final transformer block, regardless of the layer unfrozen for fine-tuning.
    \item Prompt template: \texttt{"<image>This image is"}
    \item Batch size: 1 (to avoid padding artifacts)
\end{itemize}

\paragraph{Concept model training:}
\begin{itemize}
    \item Linear regression from activations to CLIP similarity scores
    \item Concept vectors extracted as regression weights
    \item Concept weights computed from similarity matrix
\end{itemize}

\paragraph{Sensitivity computation:}
Directional derivatives were computed with respect to the completion token `` positive'', measuring how much each concept vector's direction in activation space influences the model's probability of predicting ``positive.'' This required temporarily enabling gradients for all model parameters during the backward pass.

\subsection{CLIP-Only Baseline}

As a baseline comparison, we computed CLIP scores without using any gradient information from the fine-tuned model.

The CLIP-only score was computed as:
\begin{equation}
    \text{CLIP\_score}_k = \text{Pearson}\left(\mathbf{Y}_k^{\text{test}}, \mathbf{p}^{\text{test}}\right)
\end{equation}
where $\mathbf{Y}_k^{\text{test}}$ is the vector of CLIP cosine similarities between concept $c$ and each \emph{test} image (i) in the probe set ($Y_{i,k} = \frac{\phi_v(I_i)^T \phi_t(C_k)}{\|\phi_v(I_i)\|\|\phi_t(C_k)\|}$), and $\mathbf{p}^{\text{test}}$ is the model's output probability $P(\text{positive})$ for each image in the probe/test set. This baseline tests whether simple input-output correlation (without probing internal model representations) can predict causal effects.

\subsection{Statistical Analysis}

For each feature pair and correlation condition, we recorded:
\begin{itemize}
    \item VCR sensitivity (mean across test images) for concepts A and B
    \item CLIP correlation scores for concepts A and B  
    \item Interventional effects for features A and B
\end{itemize}

Overall correlation between VCR sensitivity and interventional effects was computed using Pearson's $r$ across all 800 data points (8 feature pairs $\times$ 5 correlations $\times$ 2 features $\times$ 10 seeds). Per-feature-pair correlations were computed separately to assess consistency across visual domains. Statistical significance was assessed using scipy's \texttt{stats.pearsonr} function.

\subsection{Computational Resources}

All experiments were conducted on using 2x NVIDIA 4090 GPUs.

\subsection{Supplemental Experiments}

In addition to the figures in the main text, which show the results of the synthetic data experiments for OpenFlamingo-4B models with fine-tuning done on the parameters of the last transformer block, we repeated the experiment for (1) OpenFlamingo-4B models with fine-tuning done on the parameters of the fourth-from-last transformer block (see \figureref{fig:synthetic-of4b-4th}), (2) OpenFlamingo-3B-Instruct models with fine-tuning done on the parameters of the last transformer block (see \figureref{fig:synthetic-of3bi-last}), and (3) OpenFlamingo-3B-Instruct models with fine-tuning done on the parameters of the fourth-from-last transformer block (see \figureref{fig:synthetic-of3bi-4th}).

For all of these experiments, the VCR score is compared to the CLIP-alone score. We see that while both tend to correlate well, the correlation VLM/CLIP-alone score tends to correlate slightly better with ground truth. This is an expected result, as in this experiment the training data and testing data have similar distributions, and so the CLIP score (which doesn't need to learn a concept activation vector from a small number of samples) will tend to be more highly correlated and less susceptible to noise. This comes with the drawback, however, that it is not learning a real causal dependency of the model, and will be subject to arbitrarily bad interpretability performance if the training and probing distributions are different, as demonstrated in the next section of experiments.

\begin{figure}[]
 % Caption and label go in the first argument and the figure contents
 % go in the second argument
\floatconts
  {fig:synthetic-of4b-last}
  {\caption{The results shown in main text Figure \figureref{fig:vcr-not-fooled}, including comparison to CLIP-alone score.}}
  {\includegraphics[width=\linewidth]{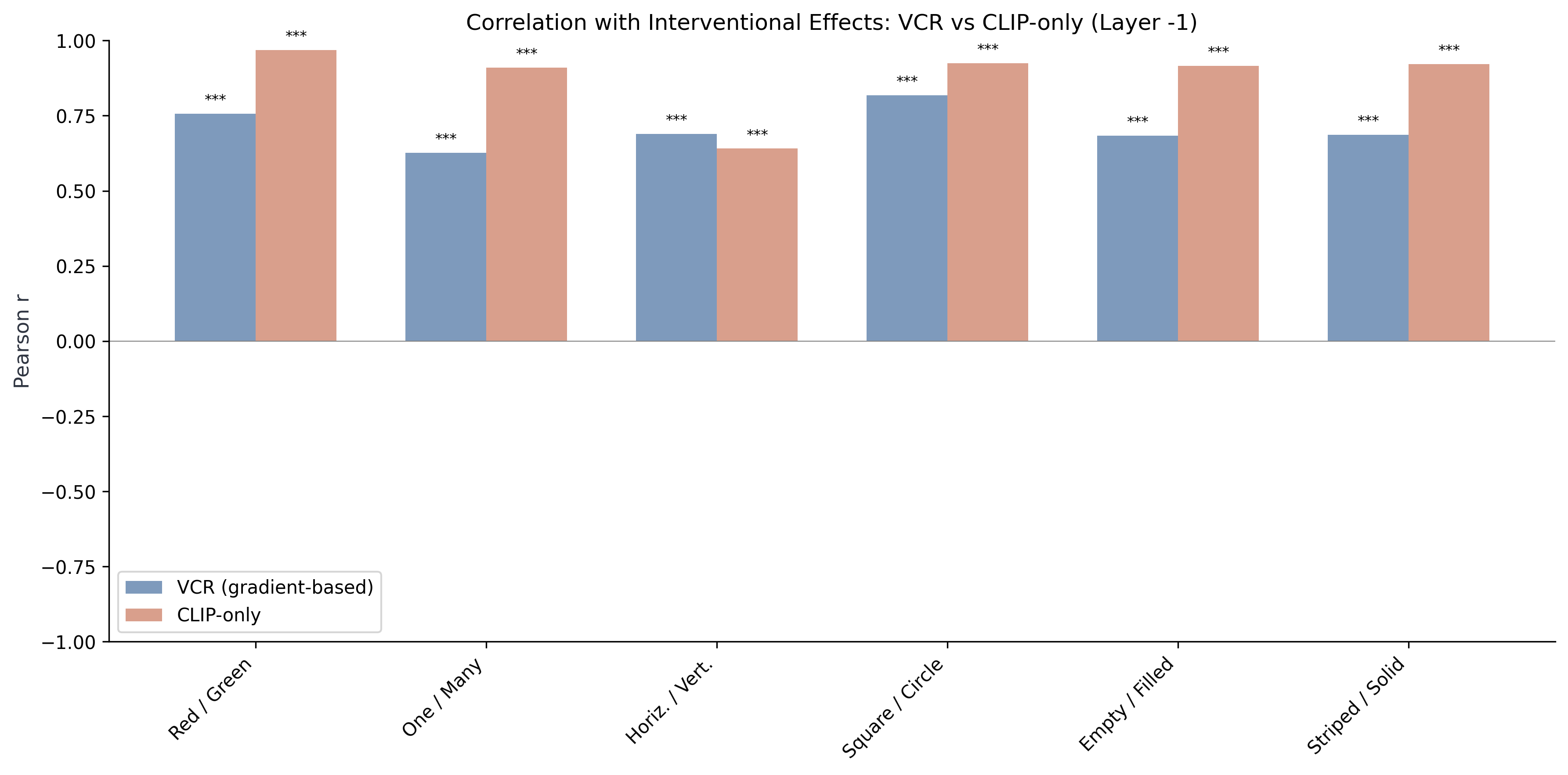}}
\end{figure}

\begin{figure}[]
 % Caption and label go in the first argument and the figure contents
 % go in the second argument
\floatconts
  {fig:synthetic-of4b-4th}
  {\caption{The results shown in main text Figure \figureref{fig:vcr-not-fooled} replicate when using OpenFlamingo-4B models with the fine-tuning done on the fourth-from-last transformer block.}}
  {\includegraphics[width=\linewidth]{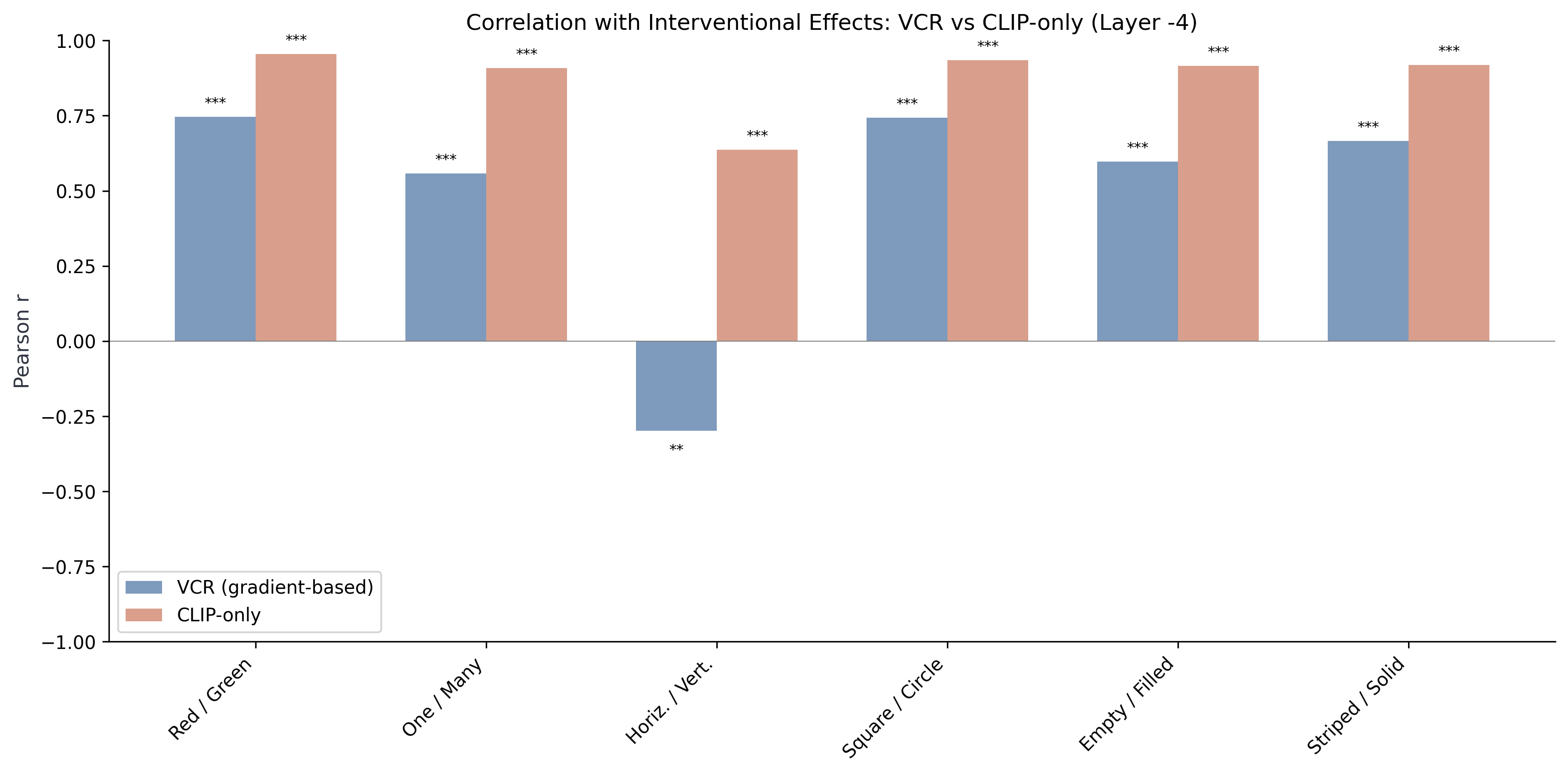}}
\end{figure}

\begin{figure}[]
 % Caption and label go in the first argument and the figure contents
 % go in the second argument
\floatconts
  {fig:synthetic-of3bi-last}
  {\caption{The results shown in main text Figure \figureref{fig:vcr-not-fooled} replicate when using OpenFlamingo-3B-Instruct models with the fine-tuning done on the last transformer block.}}
  {\includegraphics[width=\linewidth]{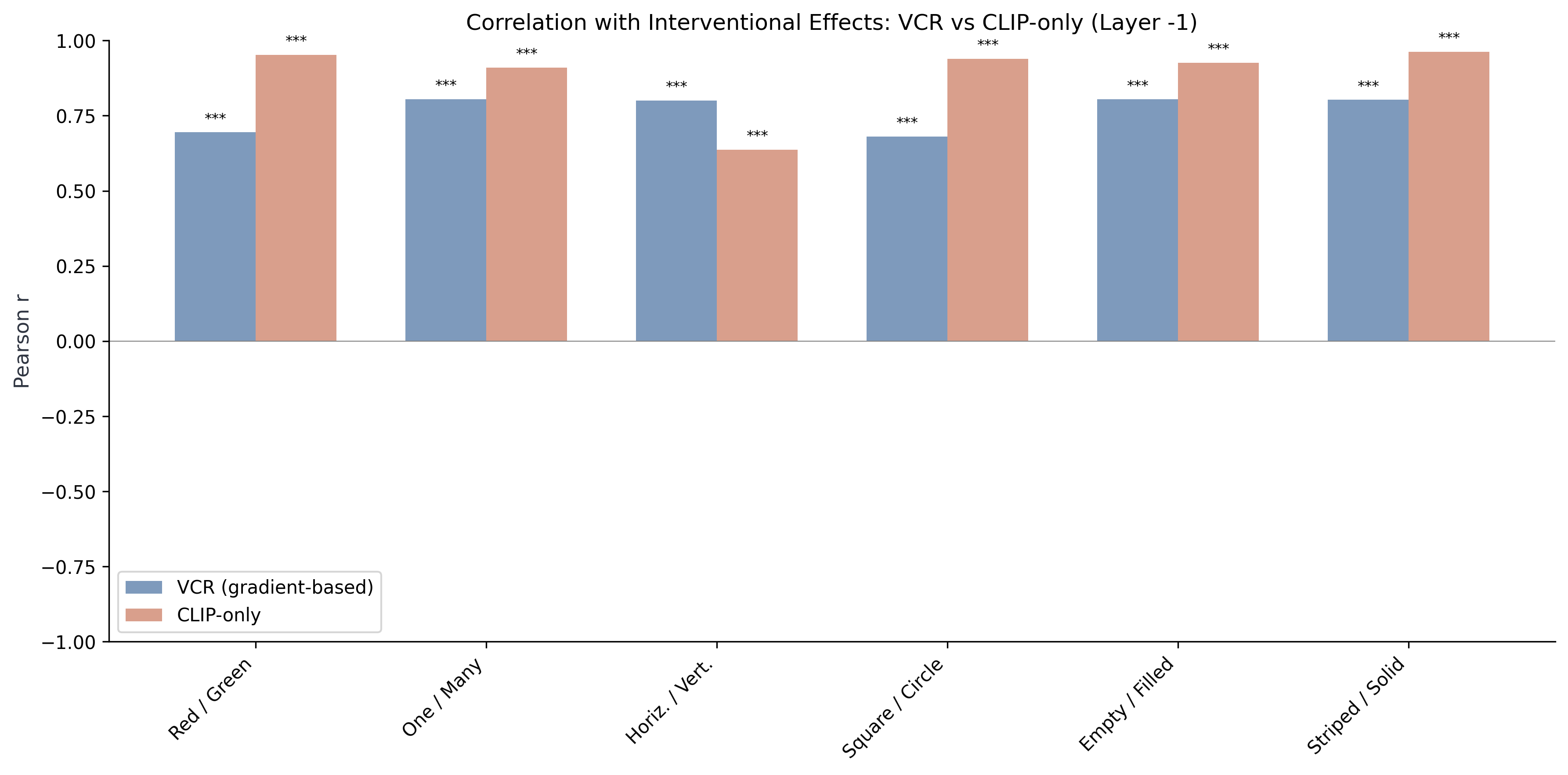}}
\end{figure}

\begin{figure}[]
 % Caption and label go in the first argument and the figure contents
 % go in the second argument
\floatconts
  {fig:synthetic-of3bi-4th}
  {\caption{The results shown in main text Figure \figureref{fig:vcr-not-fooled} replicate when using OpenFlamingo-3B-Instruct models with the fine-tuning done on the fourth-from-last transformer block.}}
  {\includegraphics[width=\linewidth]{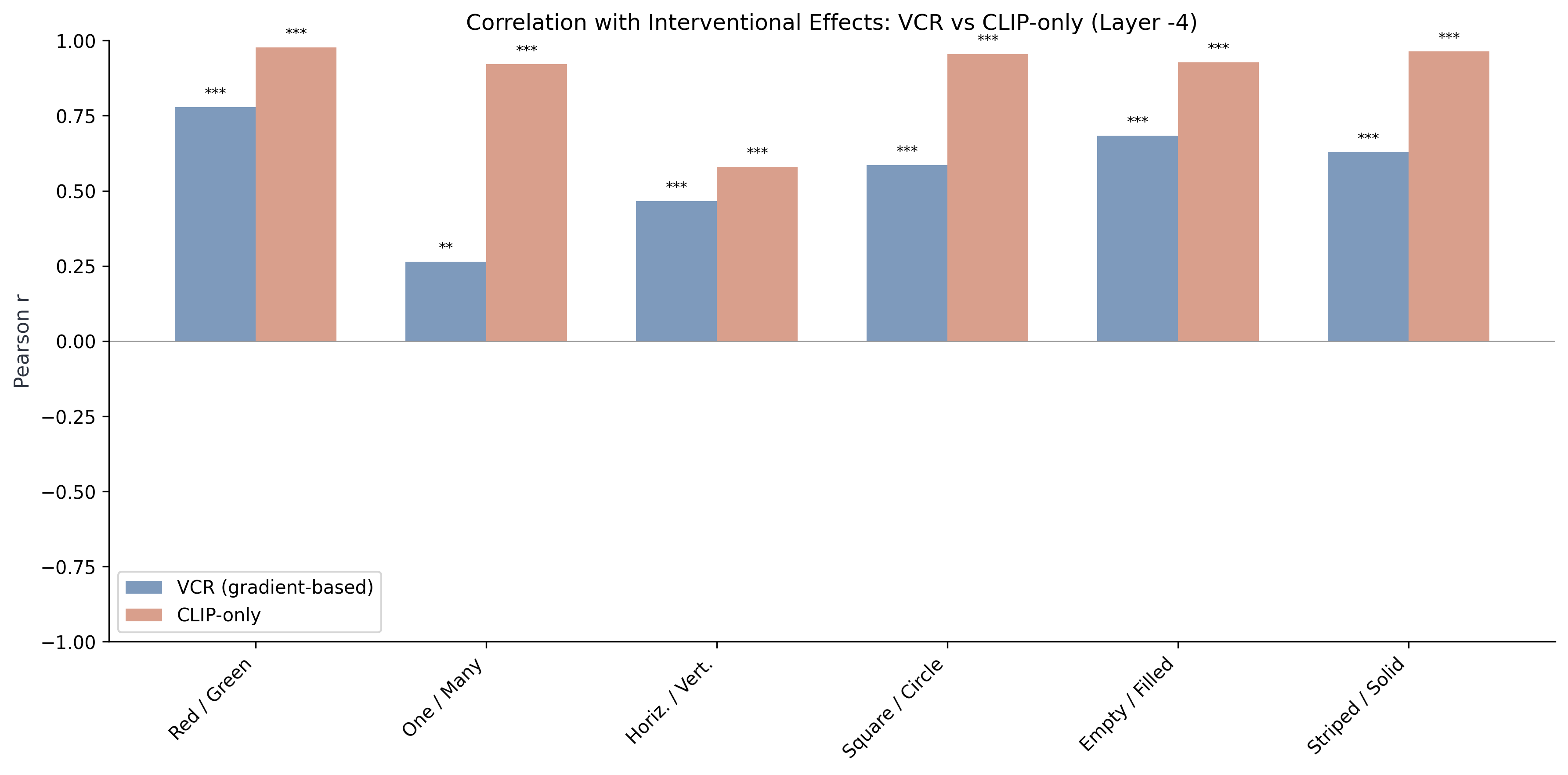}}
\end{figure}

\section{Detailed Methods for Adversarial Distribution Shift Experiment}
\label{apd:adversarial}
%%%%%%%%%%%%%%%%%%%%%%%%%%%%%%%%%%%%%%%%%%%%%%%%%%%%%%%%%%%%%%%%%%%%%%%%%%%%%%%

\subsection{Motivation and Experimental Logic}

A fundamental question for any interpretability method is whether it identifies causal features that the model is actually sensitive to perturbations in versus features that happen to correlate with model outputs in the evaluation or probe data. These can diverge when the evaluation distribution differs from training. We designed an adversarial setup that deliberately introduces such divergence: a ``spurious'' feature that negatively predicts the label during training (so the model should learn to ignore or inversely weight it) but positively correlates with the label at test time (potentially confusing correlation-based methods).

\subsection{Adversarial Dataset Construction}

For each feature pair, we construct training and test sets with deliberately mismatched feature-label correlations:

\paragraph{Training set ($N_{\text{train}} = 400$):} Feature A (reliable): Correlation $\rho_A = +1.0$ with the positive label. Feature A is present in 100\% of positive examples and 0\% of negative examples. Feature B (spurious): Correlation $\rho_B = -0.8$ with the positive label. Feature B is present in only 10\% of positive examples but 90\% of negative examples. The model learns that feature A strongly predicts ``positive'' while feature B predicts ``negative.''

\paragraph{Test (probe) set ($N_{\text{test}} = 200$):}
Feature A: Remains perfectly predictive. All positive examples have $\texttt{has\_a} = \text{True}$; all negative examples have $\texttt{has\_a} = \text{False}$. Feature B: Now \emph{positively} correlated. 80\% of positive examples have feature B present; only 20\% of negative examples have feature B. This creates a distribution shift where: $\text{Corr}(B, \text{label})_{\text{train}} = -0.8$, but $\quad \text{Corr}(B, \text{label})_{\text{test}} \approx +0.6$.

\subsection{Feature Pairs and Visual Properties}

We used six of eight feature pairs as the previously described correlation-varying experiment (with the two positional features omitted because they were not well detected in the non-adversarial setting), with identical image generation procedures. For the adversarial experiment:
\begin{itemize}
    \item Feature A (first in each pair): designated as ``reliable''
    \item Feature B (second in each pair): designated as ``spurious''
\end{itemize}

The pairs were: red/green, one/many, horizontal/vertical, square/circle, empty/filled, and striped/solid.

\subsection{Model Training Configuration}

Training followed the same protocol as the correlation-varying experiment:
\begin{itemize}
    \item \textbf{Model}: OpenFlamingo-3B-Instruct or OpenFlamingo-4B
    \item \textbf{Fine-tuning}: Last transformer block only or 4th-from-last transformer block only
    \item \textbf{Optimizer}: AdamW, $\eta = 10^{-4}$
    \item \textbf{Batch size}: 8
    \item \textbf{Epochs}: 5
    \item \textbf{Bootstrap replicates}: 10 per feature pair
\end{itemize}

\subsection{Evaluation Metrics}

For each method (VCR and CLIP-only), we computed the correlation between the score and the true interventional effect, as well as the sign-concordance between the score and interventional effect, which we defined as the fraction of ``reliable'' features with predicted positive effect, and the fraction of ``spurious'' features with predicted negative effect.

\subsection{CLIP-Only Baseline Computation}

For the adversarial experiment, the CLIP-only score was computed as:
\begin{equation}
    \text{CLIP\_score}_k = \text{Pearson}\left(\mathbf{Y}_k^{\text{test}}, \mathbf{p}^{\text{test}}\right)
\end{equation}
where $\mathbf{Y}_k^{\text{test}}$ is the vector of CLIP cosine similarities between concept $c$ and each \emph{test} image (i) in the probe set ($Y_{i,k} = \frac{\phi_v(I_i)^T \phi_t(C_k)}{\|\phi_v(I_i)\|\|\phi_t(C_k)\|}$), and $\mathbf{p}^{\text{test}}$ is the model's output probability $P(\text{positive})$ for each image in the probe/test set.

Critically, this baseline uses only test-set statistics. Since feature B is positively correlated with the label in the test set, and the model outputs are also correlated with the label (because the model performs well), CLIP-only will observe a positive correlation for feature B even though the model learned to ignore or inversely weight it.

\subsection{Supplemental Experiments}

In addition to the figure in the main text, which shows the results of the experiment for OpenFlamingo-4B models with fine-tuning done on the parameters of the last transformer block, we repeated the experiment for (1) OpenFlamingo-4B models with fine-tuning done on the parameters of the fourth-from-last transformer block (see \figureref{fig:adversarial-of4b-4th}), (2) OpenFlamingo-3B-Instruct models with fine-tuning done on the parameters of the last transformer block (see \figureref{fig:adversarial-of3bi-last}), and (3) OpenFlamingo-3B-Instruct models with fine-tuning done on the parameters of the fourth-from-last transformer block (see \figureref{fig:adversarial-of3bi-4th}).

\begin{figure}[]
 % Caption and label go in the first argument and the figure contents
 % go in the second argument
\floatconts
  {fig:adversarial-of4b-4th}
  {\caption{The results shown in main text Figure \figureref{fig:vcr-not-fooled} replicate when using OpenFlamingo-4B models with the fine-tuning done on the fourth-from-last transformer block.}}
  {\includegraphics[width=\linewidth]{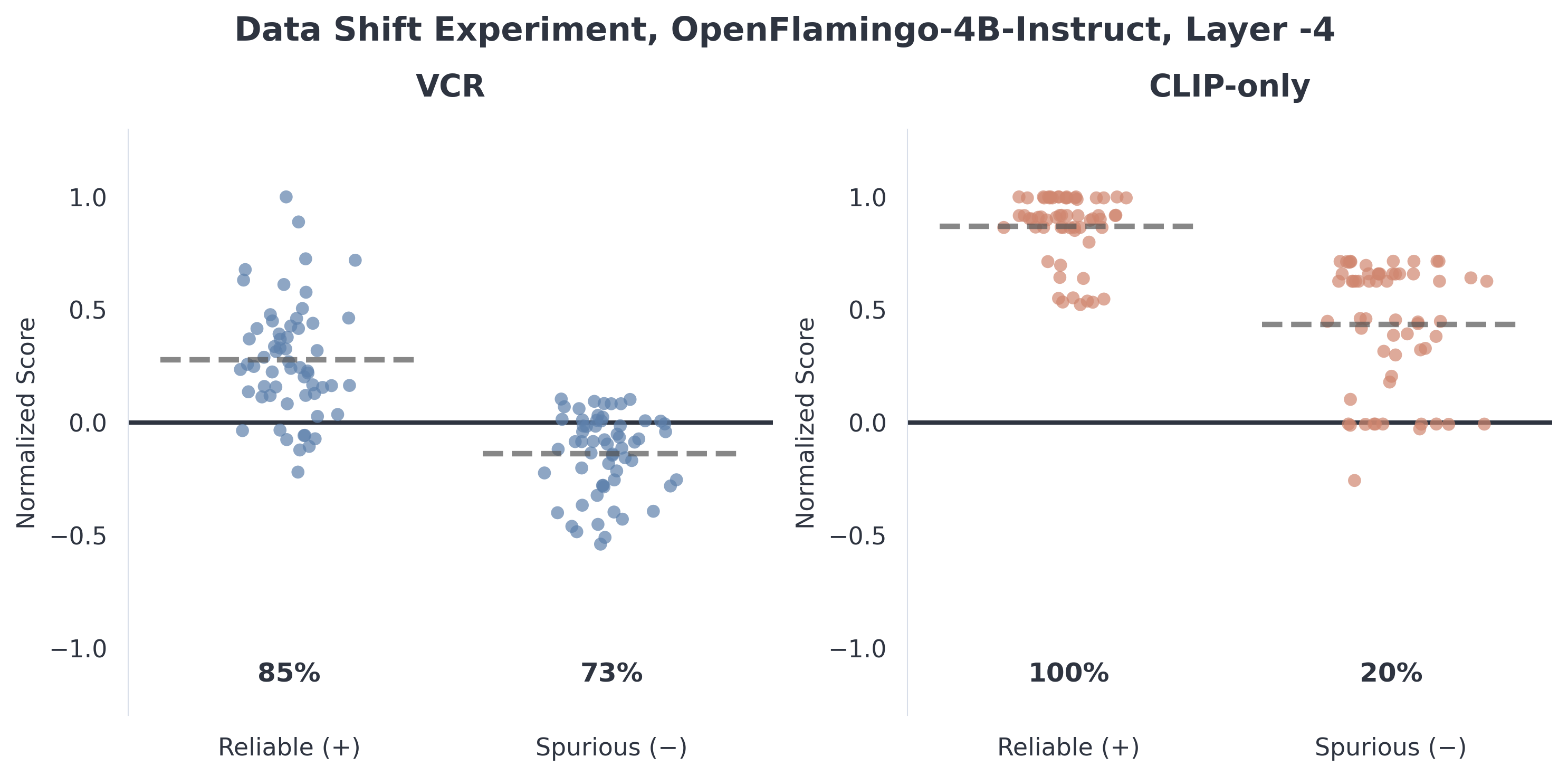}}
\end{figure}

\begin{figure}[]
 % Caption and label go in the first argument and the figure contents
 % go in the second argument
\floatconts
  {fig:adversarial-of3bi-last}
  {\caption{The results shown in main text Figure \figureref{fig:vcr-not-fooled} replicate when using OpenFlamingo-3B-Instruct models with the fine-tuning done on the last transformer block.}}
  {\includegraphics[width=\linewidth]{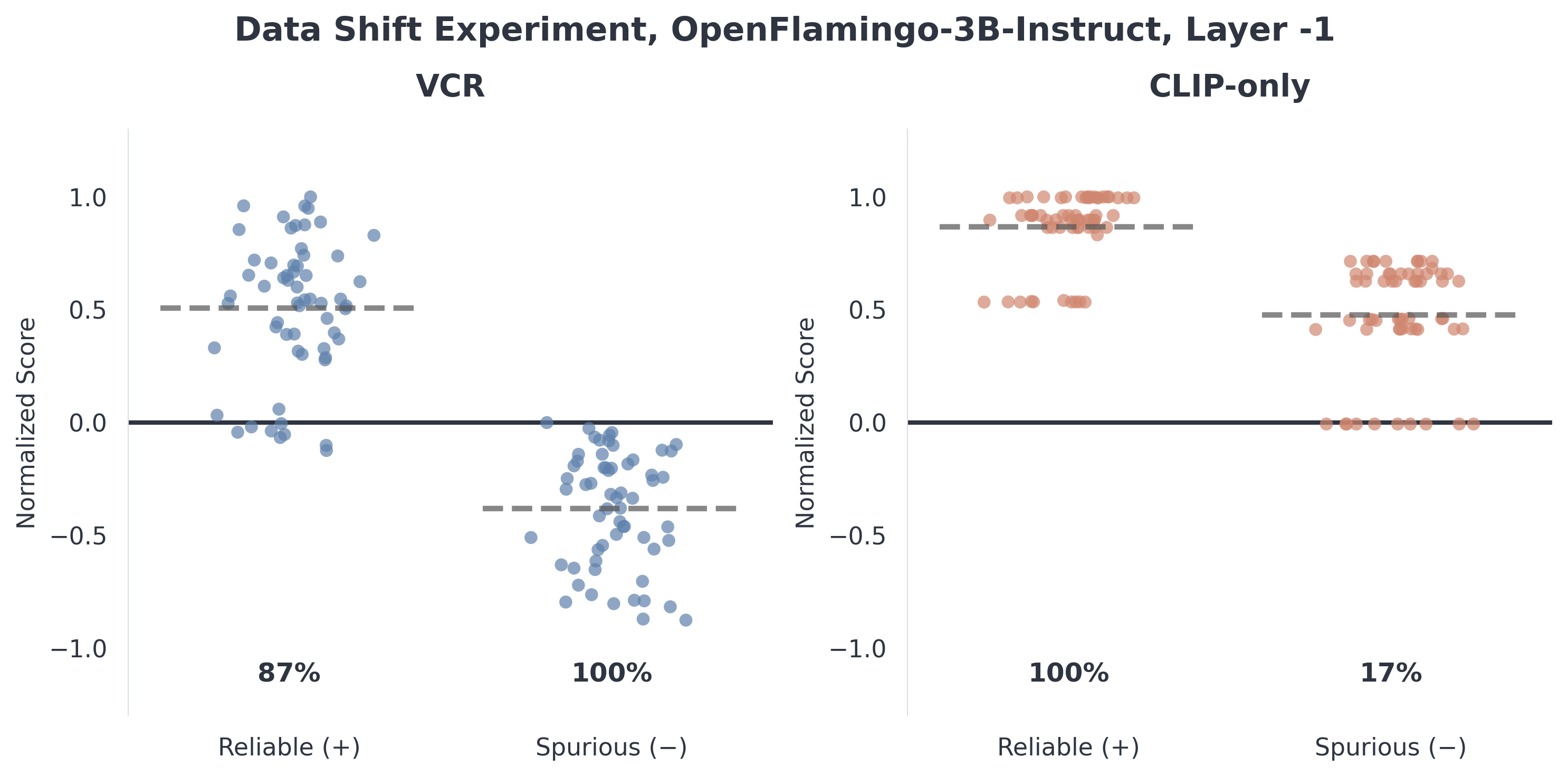}}
\end{figure}

\begin{figure}[]
 % Caption and label go in the first argument and the figure contents
 % go in the second argument
\floatconts
  {fig:adversarial-of3bi-4th}
  {\caption{The results shown in main text Figure \figureref{fig:vcr-not-fooled} replicate when using OpenFlamingo-3B-Instruct models with the fine-tuning done on the fourth-from-last transformer block.}}
  {\includegraphics[width=\linewidth]{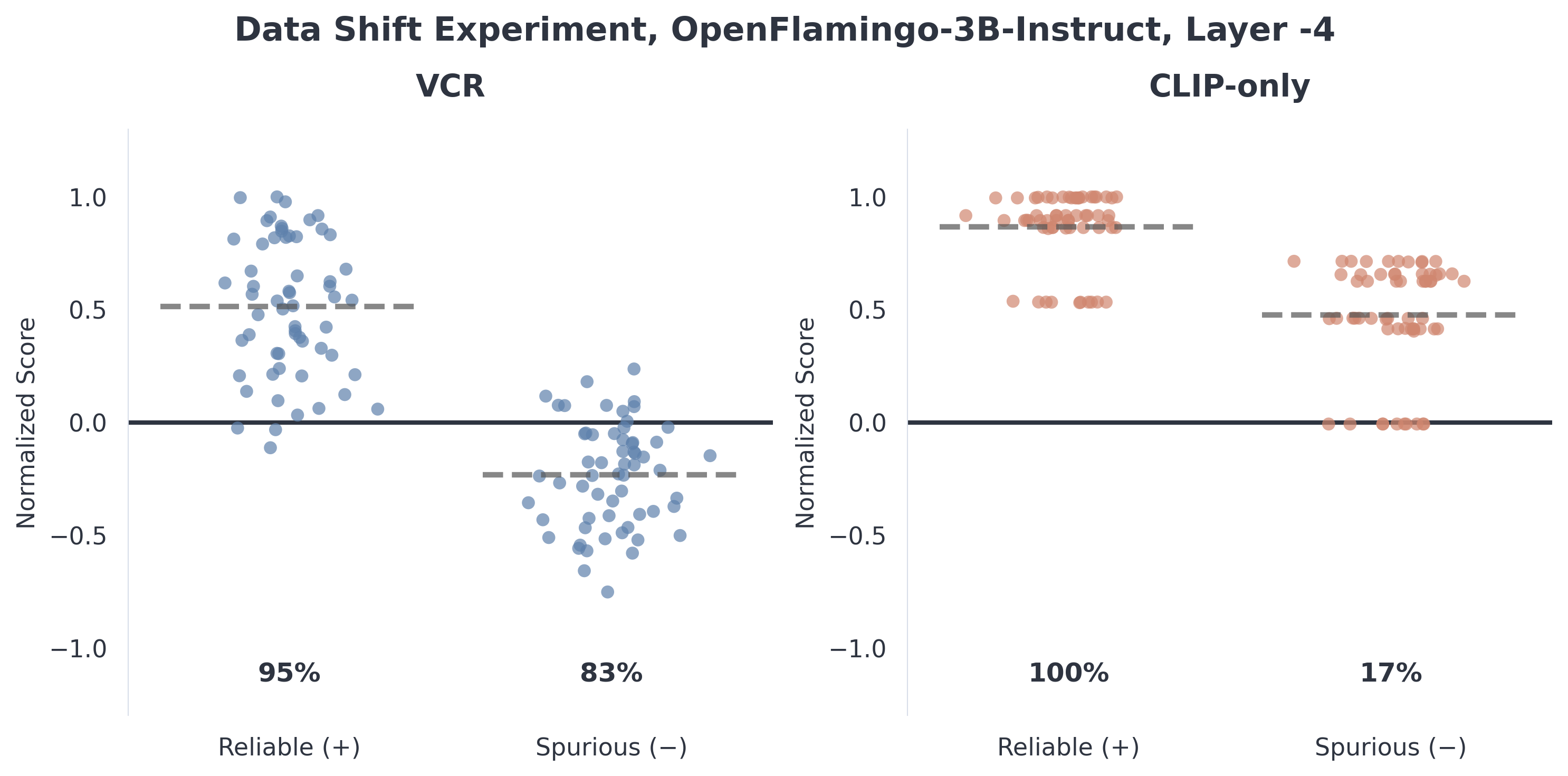}}
\end{figure}

\section{Manual Intervention Experiments}
\label{sec:manual-intervention}

To validate whether VCR correctly identifies visual features that causally influence model predictions, we designed a series of controlled intervention experiments. These experiments introduce specific visual perturbations to synthetic skin images and measure the resulting changes in model behavior, allowing us to test whether features identified by VCR as important genuinely affect classification decisions.

\subsection{Experimental Design}

We conducted three distinct intervention experiments, each targeting a different type of visual feature: (1) colored dot artifacts (corresponding to the skin markings made by dermatologists prior to biopsies), (2) scalp backgrounds, and (3) facial backgrounds. All experiments followed a common framework: synthetic skin images were modified with controlled visual perturbations, and the model's predicted probability of malignancy was measured before and after intervention.

\paragraph{Synthetic Image Generation.}
We used synthetic skin patch images generated to represent two Fitzpatrick skin type (FST) groups: FST I-II (lighter skin tones) and FST V-VI (darker skin tones). These synthetic images contain no actual lesions, providing a neutral baseline against which to measure the effect of spurious visual features on malignancy predictions. These were generated by prompting Gemini-3 Image Creation mode with the following prompts: ``Create a close up of Fitzpatrick Skin Type \texttt{insert type here} skin, as would be seen for a realistic clinical dermatology image in a dermatology textbook.''

\paragraph{Model and Inference.}
We evaluated OpenFlamingo models (3B-Instruct for the colored dot experiment; 4B for the scalp and face experiments). For each test image, we constructed prompts containing demonstration examples sampled from the Diverse Dermatology Images (DDI) dataset with ground-truth labels. Zero-shot prompts followed the format:

\begin{verbatim}
Based on the image, this lesion is benign.
<|endofchunk|>
Based on the image, this lesion is malignant.
<|endofchunk|>
<i>Based on the image, this lesion is
\end{verbatim}

While the ICL prompt followed the format:

\begin{verbatim}
Based on the image, this lesion is benign.
<|endofchunk|>
Based on the image, this lesion is malignant.
<|endofchunk|>
<i>Based on the image, this lesion is {lab}.
<|endofchunk|>
...
<i>Based on the image, this lesion is
\end{verbatim}

\noindent where \texttt{<i>} indicates the image tag, \texttt{{lab}} is replaced with the demonstration label, and demonstration images and labels preceded the test/query image. We measured the length-normalized log-probability of the token ``malignant'' as our primary outcome metric, computed via:
\begin{equation}
    \frac{1}{|\text{malignant}|} \sum_{t} \log P(x_t | x_{<t}, \text{context})
\end{equation}

\paragraph{Demonstration Example Selection.}
Demonstration examples were sampled from a held-out subset of the DDI dataset. For each experiment, we selected 4 malignant and 12 benign examples (a 1:3 ratio reflecting class imbalance in clinical settings), balanced across FST I-II and FST V-VI skin types. Examples were shuffled to prevent ordering effects.

\paragraph{Data Augmentation.}
To increase statistical power and reduce variance from specific image characteristics, each synthetic test image was augmented 10 times using random transformations: rotation (up to $\pm 15°$), translation (up to 10\% of image dimensions), and random resized crops (scale factor 0.85--1.0) to a final resolution of $224 \times 224$ pixels.

\subsection{Colored Dot Experiment}

This experiment tested whether the model associates small blue/purple dots with malignancy predictions, a potential source of spurious correlation in clinical datasets where dermatologist-applied skin markings may correlate with lesion severity.

\paragraph{Intervention.}
For each synthetic skin image, we added four circular dots of radius 30 pixels at random locations. Dots were colored with an identical purple hue (RGB: 60, 34, 112) for both FST groups.

\paragraph{Conditions.}
We evaluated four conditions in a $2 \times 2$ factorial design: skin type (FST I-II vs.\ FST V-VI) $\times$ dots present (yes vs.\ no). Both zero-shot and ICL settings were tested.

\paragraph{Analysis.}
For each condition, we computed the mean change in log-probability of malignancy:
\begin{equation}
    \Delta = \overline{\text{logprob}(\text{malignant})}_{\text{dots}} - \overline{\text{logprob}(\text{malignant})}_{\text{no dots}}
\end{equation}
Standard errors were estimated via bootstrap resampling of the difference in means.

\subsection{Facial Background Experiment}

This experiment tested whether the presence of patient faces in the image background---a feature that should be irrelevant to dermatological diagnosis---affects malignancy predictions.

\paragraph{Intervention.}
We compared: (1) baseline synthetic skin patches with neutral backgrounds, and (2) synthetic skin patches composited onto images of human faces. Both sets used FST I-II skin tones, and the face images were selected to match the skin tone of the lesion patches.

\paragraph{Analysis.}
Identical to the scalp experiment: we computed the difference in mean log-probability and assessed significance via independent-samples $t$-test.

\subsection{Scalp Background Experiment}
\label{apd:scalp}

This experiment tested whether anatomical context (specifically, scalp backgrounds with visible hair) influences malignancy predictions independently of lesion characteristics.

\paragraph{Intervention.}
We compared two image sets: (1) baseline synthetic skin patches with neutral backgrounds, and (2) synthetic skin patches composited onto scalp backgrounds showing hair patterns. Both sets used FST I-II skin tones.

\paragraph{Analysis.}
We computed the difference in mean log-probability of malignancy between scalp-background and neutral-background conditions. Statistical significance was assessed using an independent-samples $t$-test:
\begin{equation}
    t = \frac{\bar{x}_{\text{scalp}} - \bar{x}_{\text{baseline}}}{s_p \sqrt{2/n}}
\end{equation}
where $s_p$ is the pooled standard deviation.

\subsection{Implementation Details}

All experiments were implemented in Python using PyTorch and the Hugging Face Transformers library. Image augmentations were applied using \texttt{torchvision.transforms}. Model inference was performed on a single NVIDIA GPU. Random seeds were fixed at 42 for reproducibility across all experiments.

For each experiment, raw log-probabilities were saved to CSV files for subsequent analysis. Visualization was performed using Matplotlib and Seaborn with a consistent visual style across all figures.

\section{Computational Efficiency}

\begin{figure}[ht]
\begin{center}
%\framebox[4.0in]{$\;$}
\includegraphics[width=\linewidth]{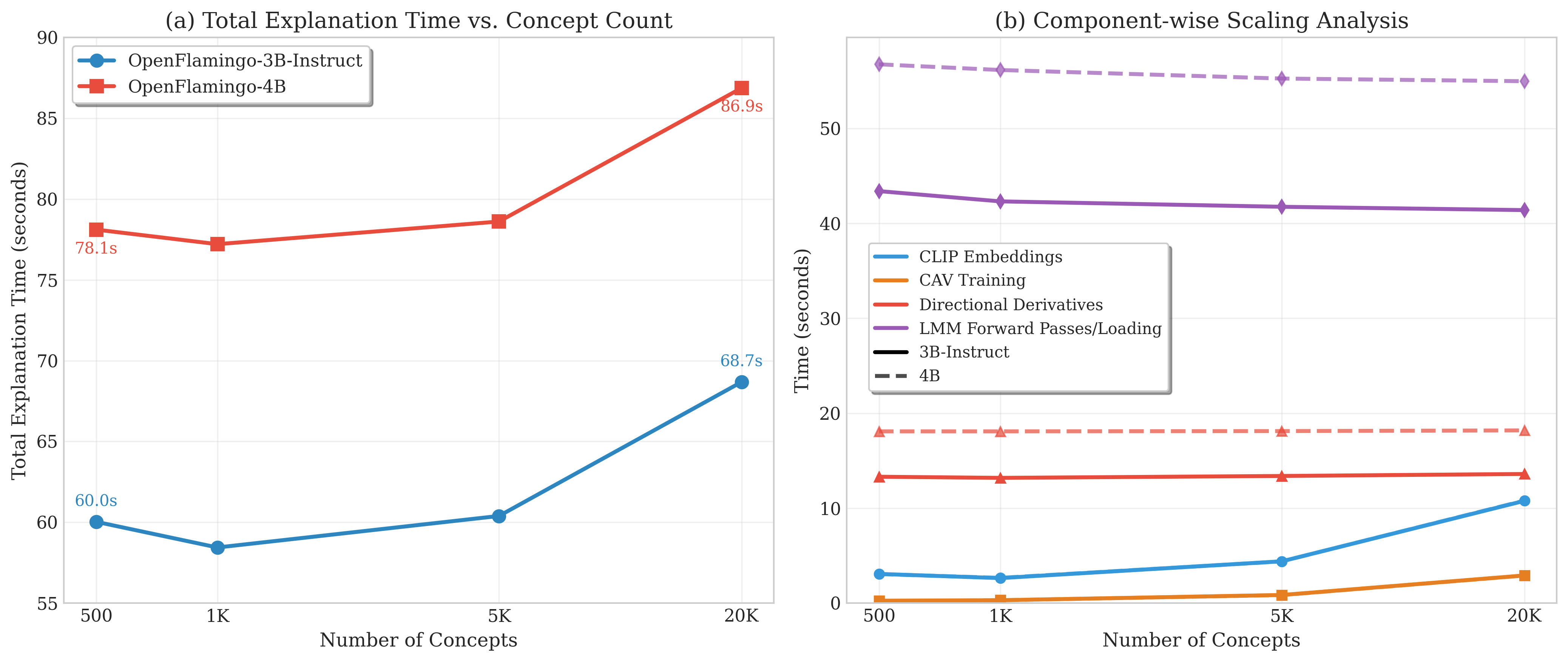}
\end{center}
\caption{Wall-clock time to generate explanations for the last layer of two OpenFlamingo models, as the number of concept explanations generated is varied.}
\label{fig:concept_scaling}
\end{figure}

We also analyzed the wall-clock time of VCR (see \figureref{fig:concept_scaling}A). Using the 328 train samples from the DDI dataset (half of the dataset) with the zero-shot prompt, we found that it took approximately 60 to 90 seconds on a single NVIDIA GeForce RTX 4090 GPU to generate between 500 and 20K concept explanations for a single layer. Importantly, we see that adding additional concepts did not significantly impact the overall time required, adding \textit{only 8 seconds} to generate 20K vs 500 concept explanations (Fig. \ref{fig:concept_scaling}B). This is because the process of generating VLM embeddings and training CAVs was relatively small compared to fixed costs like loading the LMM onto the GPU, the forward passes required to extract log-probs and activations, and the backward passes required to calculate directional derivatives. We also analyze the impact of changing the probe set size, which has a much larger impact on the total time required. This effect is linear in the number of images in the probe set, as this increases the number of forward passes and backward passes of the model required. We highlight here that our approach remains so computational feasible that we are also able to use \textit{an order of magnitude} larger probe set sizes than those described in prior work such as \citet{kim2018interpretability}, where they demonstrated effective CAV training with approximately 30 example images. Our probe set sizes were similar to prior automated approaches like LG-CAV \citep{huang2024lg}.

Our timing analysis focused on five key computational components: (1) model loading, (2) CLIP embedding computation, (3) concept model training, (4) directional derivatives calculation, and (5) other processing steps including activation collection and concept weight computation. Both the concept-scaling and probe set-scaling experiments were run with a fixed random seed to ensure reproducible data splits and timing measurements.

To evaluate how explanation time scales with concept vocabulary size, we varied the number of concepts while holding other parameters constant. We tested four concept vocabulary sizes: 500, 1,000, 5,000, and 20,000 concepts. The probe set was constructed by randomly sampling training images from the DDI dataset, with a fixed train/test split ratio of 0.5 and consistent data preprocessing across all concept vocabulary sizes. We used layer 23 for OpenFlamingo-3B-Instruct and layer 31 for OpenFlamingo-4B, representing the final decoder layers in each model architecture.

To assess computational scaling with respect to training data size, we varied the probe set size while maintaining a fixed concept vocabulary of 1,000 concepts. We tested three probe set sizes: 50, 100, and 200 training samples. For each probe set size, we randomly sampled the specified number of training images from the DDI dataset, with remaining samples allocated to the test set.
This design allowed us to isolate the computational impact of increasing the number of images for which activations must be collected and gradients computed, while controlling for concept vocabulary complexity. The same model layers and processing pipeline were used as in Experiment 1.

Each experiment measured wall-clock time for individual computational components using Python's time module. We recorded the duration of each major processing step and computed total explanation time as the sum of all components. To account for potential GPU memory and caching effects, we performed cleanup operations between different parameter settings, including removal of temporary files and explicit garbage collection. The timing measurements capture the end-to-end computational cost of generating explanations for a single model layer, including all preprocessing, model inference, and post-processing steps required by our approach.

\begin{figure}[ht]
\begin{center}
%\framebox[4.0in]{$\;$}
\includegraphics[width=\linewidth]{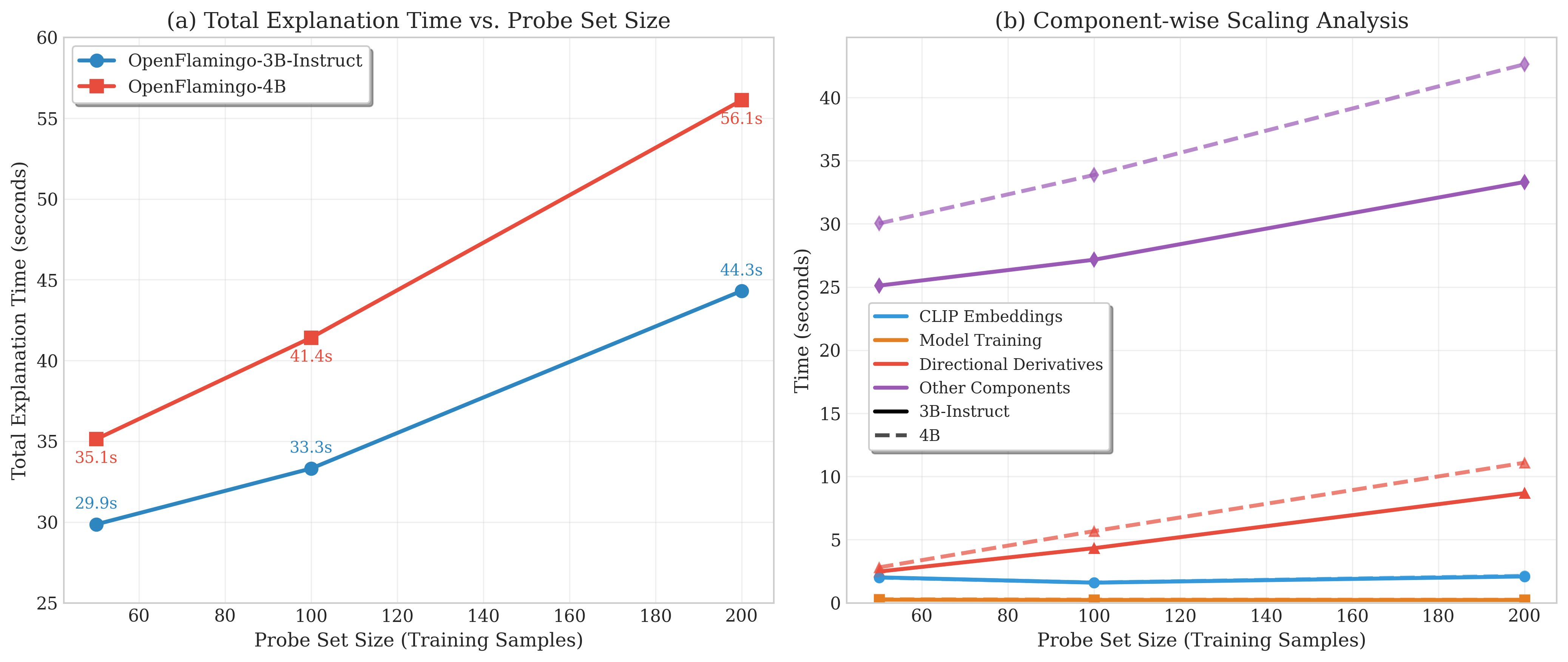}
\end{center}
\caption{Wall-clock time to generate explanations for the last layer of two OpenFlamingo models, as the number of images in the probe set is varied.}
\label{fig:probe_scaling}
\end{figure}

\section{CheXpert Chest Radiographs}\label{apd:chexpert}

To demonstrate the applicability of our VCR approach beyond dermatology, we conducted our analysis using the CheXpert chest radiograph dataset. We analyzed the OpenFlamingo-4B model, focusing on the final decoder layer (layer 31 of the GPT-NeoX language encoder). We selected 400 frontal chest radiographs, balanced between normal findings (n=200) and abnormal findings (n=200), with patient-level splitting to prevent data leakage. The dataset was divided 50/50 into training and test sets using stratified sampling by patient ID. We employed a zero-shot prompting strategy with the template ``Based on the image, this radiograph is normal/abnormal'' to elicit binary classification behavior. We used approximately 10,000 diverse concept terms from a filtered English word list. 

To ensure stability of concepts we repeated the complete VCR analysis pipeline across 25 independent random seeds (seeds 0-24), with each seed producing a different train/test split. For each seed iteration, we computed CLIP text-image similarity matrices, collected layer activations, trained linear concept models with L2 regularization, calculated concept weights from empirical variance, and computed weighted directional derivatives to rank concept sensitivity. We then performed statistical significance testing using one-sample t-tests (two-sided, null hypothesis: mean directional derivative = 0) across all seeds for each concept. To control for multiple comparisons across the approximately 10,000 concepts tested, we applied Bonferroni correction ($\alpha = 0.05 / n_{\text{concepts}}$). We identified the top 20 significantly positive concepts (highest mean directional derivatives toward ``abnormal'' predictions) and top 20 significantly negative concepts (lowest mean directional derivatives, favoring ``normal'' predictions). For visualization purposes, we used the CLIP similarity matrix from seed 0 to identify the most and least activating probe images for each significant concept, displaying 7 representative examples from each extreme using center-cropped square images at 300 DPI resolution.

\begin{figure}[t]
  \centering
  \includegraphics[width=\linewidth]{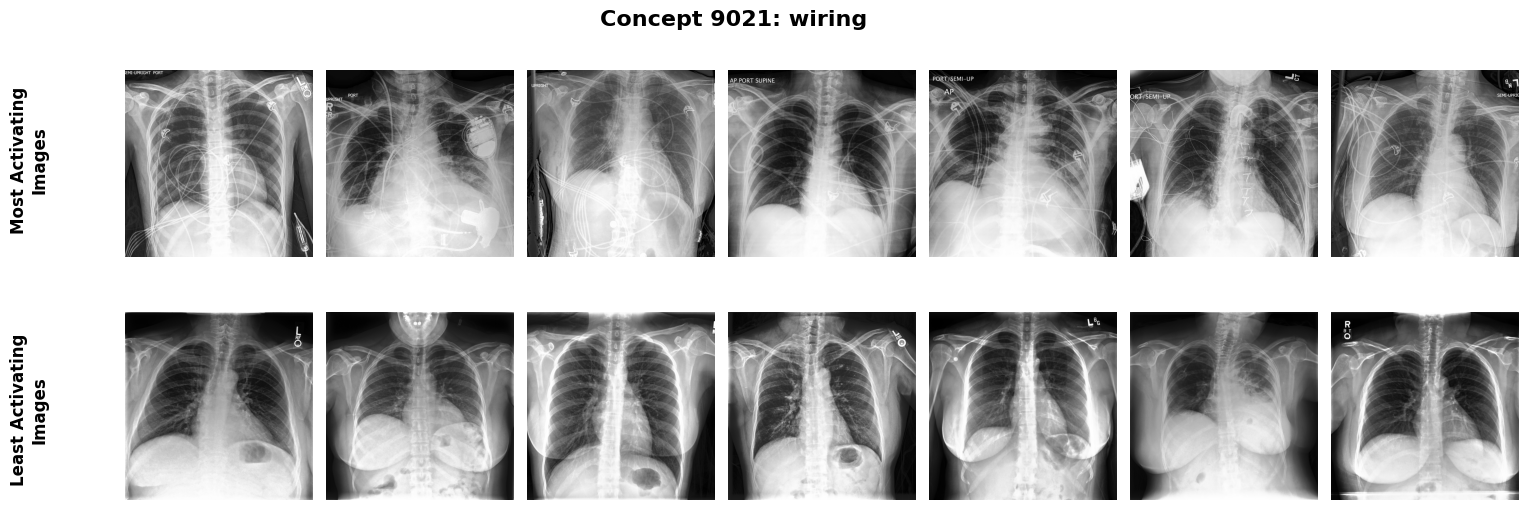}
  \caption{Most (top) and least (bottom) activating images for wiring concept.}
  \label{fig:wiring}
\end{figure}

One of the top concepts identified by this analysis is plotted in \figureref{fig:wiring}, where the top row shows the most activating images. These radiographs show a wide diversity of medical support devices. For instance, the first from left has EKG leads; the second from left has has EKG leads, a ventricular assist device, as well as a pacemaker with pacemaker leads; etc.

\section{Imagenette}\label{apd:imagenette}

\begin{figure}[t]
  \centering
  \includegraphics[width=\linewidth]{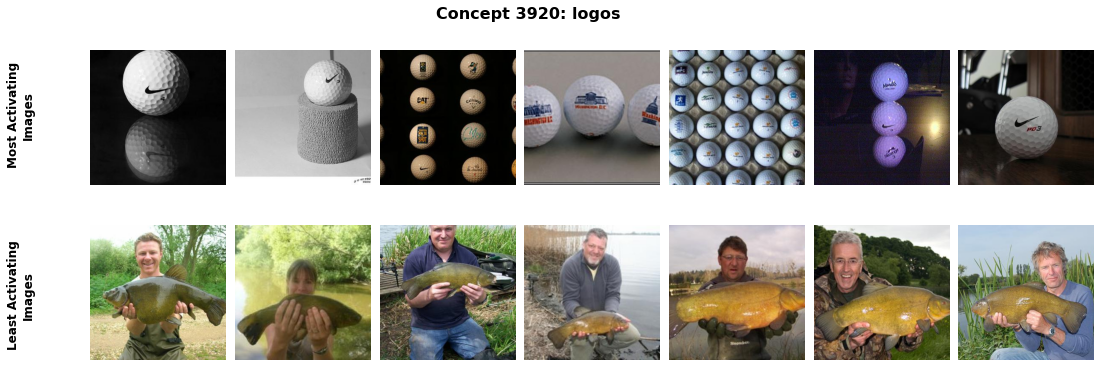}
  \caption{Most (top) and least (bottom) activating images for logos concept.}
  \label{fig:logos}
\end{figure}

As mentioned in our Discussion section, the applications demonstrated in the main text likely merely scratch the surface of potential application of our VCR approach. To demonstrate that our approach can be applied to other models, data, and task types, we generated VCR explanations for a larger, 9B parameter OpenFlamingo model. We focused on the self-attention modules within the decoder layers. We used natural images from ImageNette to test the model, specifically selecting two distinct object classes: Fish images (Class n01440764 from Imagenette) and Golf ball images (Class n03445777 from Imagenette). The dataset was split 50/50 into training and test sets using a fixed random seed (1017) to ensure reproducibility. We employed 3,000 diverse concept terms from a broad vocabulary file. We designed a few-shot learning template to elicit binary classification behavior. The resulting top concepts (see Appendix Table \ref{tab:fish_golf_concepts}) demonstrate that: (1) the model's fish concepts are semantically coherent (biological, aquatic terms), (2) Golf ball concepts relate to sports equipment and visual patterns (see Appendix Fig. \ref{fig:dotted}, which shows that a dotted visual pattern is important, and Appendix Fig. \ref{fig:logos}, which shows that the visual presence of logos in the image leads a model to predict an image is a golf ball). The model successfully learns to distinguish domain-specific features at the attention layer level, not just in the residual stream.

\begin{table}[h!]
\centering
\small
\caption{Top 25 visual concepts for Fish vs Golf Ball Classification in 9B parameter Flamingo model's 25th decoder layer self-attention module.}
\label{tab:fish_golf_concepts}
\begin{tabular}{|l|l|}
\hline
\multicolumn{2}{|c|}{\textbf{Fish Concepts}} \\
\hline
aquaculture & fisheries \\
suriname & fish \\
carp & freshwater \\
mutant & fishing \\
bass & peruvian \\
female & specimen \\
biodiversity & fishery \\
unit & discarded \\
marrow & futuna \\
syndicate & meal \\
nunavut & pike \\
sphere & guyana \\
syrian & \\
\hline
\multicolumn{2}{|c|}{\textbf{Golf Ball Concepts}} \\
\hline
putting & logos \\
tights & dotted \\
incorporates & pointing \\
plaid & stripes \\
visor & affiliations \\
pointers & overall \\
iu & acura \\
pantera & lettering \\
docs & paige \\
scottsdale & pines \\
memberships & belts \\
tapestry & rings \\
classicvacations & \\
\hline
\end{tabular}
\end{table}

\begin{figure}[t]
  \centering
  \includegraphics[width=\linewidth]{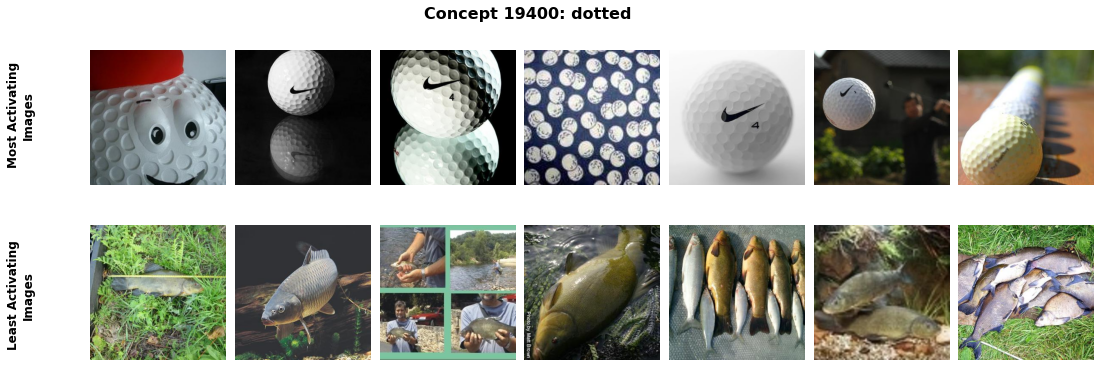}
  \caption{Most (top) and least (bottom) activating images for dotted concept.}
  \label{fig:dotted}
\end{figure}

\section{Related interpretability papers}\label{apd:rel_work}

While we do not call our approach an ``automatic interpretability'' method, because we believe this is somewhat of a misleading way to describe a method that works best with a human-in-the-loop to examine concepts and run validating experiments, our method has a lot in common with and draws from techniques that have been branded as ``automatic interpretability.'' For instance, previous papers have used VLMs for automatic interpretability. For example, CLIP-Dissect \citep{oikarinen2022clip} used CLIP to label the most important concepts for individual neurons within a neural network. \citet{kim2024transparent} first designed a specially-trained dermatology-specific foundation model that they called MONET, then used MONET to find concepts that correlate with a model's output or a model's loss, an approach they called Model Auditing with MONET (MA-MONET). 

Other approaches to automatic interpretability include \citet{zhang2024large}, where authors trained SAEs on the activations of a small LMM (LLaVA-NeXT-8B) to produce disentangled latent features, then used a larger LMM (LLaVA-OV-72B) to automatically interpret those latents by generating natural language descriptions from top-activating examples. Similarly, \citet{paulo2024automatically} build a pipeline using LLMs to automatically generate natural language explanations for SAE features from other LLMs. 

Beyond automatic interpretability, there has been substantial research into linear representations of concepts and linear probing of neural networks \citep{belinkov2022probing,park2023linear}. For example, a recent paper by \citet{rajaram2025line} trained linear probes on residual stream activations of LMMs and demonstrated that image features are represented linearly, become more multimodal in deeper layers, and can be manipulated to control model behavior. \citet{kim2018interpretability} introduced Testing with Concept Activation Vectors (TCAV), a method to quantify how much user-defined, human-interpretable concepts influence a model’s predictions by measuring directional derivatives along concept vectors in activation space. \citet{zou2023representation} develop RepEng, an approach for testing LLMs' sensitivity to particular textual concepts, and steering their use of those concepts.

Finally, there has been extensive work into interpretability for skin images. For instance, \citet{degrave2025auditing} and \citet{gadgil2025dream} both propose methods involving generation of counterfactual images to investigate various aspects/behaviors of clinical dermatology models. \citet{pmlr-v287-jin25b} investigate a small number of artifactual confounders for dermoscopy classifiers using denoising diffusion models. \cite{yan2023towards} make use of CAVs in experiments on building interpretable/steerable dermatology models, and \cite{nicolson2024explaining} demonstrate some potential short-comings of their method.

\end{document}